\documentclass[11pt]{article}
\usepackage[preprint]{acl}
\usepackage{times}
\usepackage{latexsym}
\usepackage[T1]{fontenc}
\usepackage[utf8]{inputenc}
\usepackage{microtype}
\usepackage{inconsolata}
\usepackage{graphicx}
\usepackage{amsmath,amssymb,amsthm}
\usepackage{booktabs}
\usepackage{graphicx}
\usepackage{xcolor}
\usepackage{hyperref}
\usepackage{enumitem}
\usepackage[normalem]{ulem} 

\newtheorem{definition}{Definition}

\theoremstyle{remark}

\usepackage{tabularx}
\usepackage{multirow}
\usepackage{multirow, makecell, caption, xcolor, array}
\usepackage[table]{xcolor}
\usepackage{soul}
\usepackage{siunitx}
\usepackage{float}
\usepackage{placeins}
\usepackage{subcaption}   
\usepackage{algorithm}
\usepackage{algpseudocode}
\renewcommand{\arraystretch}{1.18}

\newcommand{\E}{\mathbb{E}}

\definecolor{directbg}{RGB}{255,243,224}
\definecolor{divbg}{RGB}{224,238,255}
\definecolor{kwbg}{RGB}{224,248,232}

\newcolumntype{L}{>{\raggedright\arraybackslash\small}p{0.305\textwidth}}

\title{Where You Inject Diversity Matters:\\ A Unified Framework for Diverse Generation}

\author{Cheng Zhang \\
   UNC Chapel Hill\\
  \texttt{zch@cs.unc.edu} \\\And
  Rui Xin \\
  University of Washington\\
  \texttt{rx31@cs.washington.edu} \\\And
  Chudi Zhong \\
  UNC Chapel Hill\\
  \texttt{chudi@unc.edu} \\
  }

\begin{document}
\maketitle
\begin{abstract}

Open-ended generation tasks often require a set of meaningfully different outputs, yet large language models often produce similar generations. Existing test-time diversity methods operate at different stages of generation with varying effectiveness, but it remains unclear what design choices lead to meaningful diversity in the output. We introduce a framework that characterizes test-time diverse generation methods by the diversity source introduced during generation and provide a transmission score for measuring how effectively variation in the source reaches the final output. Guided by this framework, we propose fully automated specification-level generation methods that first generate diverse intermediate specifications and then condition on them to produce final responses. Across five open-ended tasks and four backbone models, specification-level injection improves output diversity over test-time baselines while maintaining comparable quality. Our analysis shows that successful diversity injection depends on both the diversity of the sources and their transmission to the output, highlighting source design and source-to-output realization as two key levers for building more diverse generation systems.

\end{abstract}

\section{Introduction}
\label{sec:intro}

Large language models (LLMs) are increasingly used for open-ended generation tasks such as creative writing, brainstorming, argumentation, and ideation. In these settings, users often need a set of meaningfully different outputs rather than a single best answer. However, when prompted repeatedly with the same request, LLMs tend to converge on similar high-level patterns and recurring ideas, producing outputs that differ in wording but not in substance \citep{sorensen2024position, jiang2026artificial}. This limits the practical value of generating multiple outputs and risks homogenizing the creative input available to users.
 
A growing body of test-time methods attempts to address this problem by increasing decoding randomness through temperature scaling or modified sampling strategies \citep{fan2018hierarchical, holtzman2020curious, minh2025minp}, introducing per-output random strings or noise tokens during generation \citep{misaki2026string, agrawal2026addressing}, or steering decoding with diversity-promoting modules \citep{ruan2025g2}. These methods operate at different stages of the generation process and inject different forms of variation, raising a natural question: what kind of diversity source is more effective for producing diverse final outputs?

In this work, we propose a \textbf{Diversity Injection Framework}, a unified lens for studying test-time diverse generation. We formulate generation as $y_i = G(x, z_i)$, where $x$ is the input prompt and $z_i$ is an explicit diversity source that varies across outputs. Methods differ in whether $z_i$ exists and what information it carries. We categorize methods into three levels: \emph{no injection} (Level~0), where no per-output $z_i$ exists; \emph{surface-level injection} (Level~1), where $z_i$ carries no semantic content about the output; and \emph{specification-level injection} (Level~2), where $z_i$ provides a detailed generative configuration. We introduce an entropy decomposition of output diversity into \emph{injected diversity} (how diverse the sources $z_i$ are), \emph{information loss} (diversity in $z$ not realized in the output), and \emph{residual randomness} (variation from decoding). From this decomposition we derive a \emph{transmission score} that quantifies what fraction of injected diversity reaches the final output.

Analysis through the framework reveals that existing fully automated diversity methods remain at Level~0 or Level~1, relying entirely on decoding randomness or injecting diversity sources that carry little semantic content. Level~2 injection has not yet been fully explored for promoting diverse generation. This gap motivates us to ask: \emph{can specification-level diversity injection improve output diversity across general open-ended tasks?}

We propose fully automated methods that instantiate Level~2 injection and are applicable to broad open-ended generation tasks. Given a prompt, the model first generates multiple diverse specifications, each describing a distinct generative configuration along dimensions such as perspective, tone, and thematic focus. The model then conditions on each specification to produce a final output. The entire process requires no human intervention and no task-specific engineering: the same pipeline applies to creative writing, argumentation, poetry, and brainstorming alike. 

In summary, we make three contributions: (1) a unified framework that characterizes test-time diverse generation methods by the nature of their diversity source, and an entropy-based transmission score that quantifies how effectively injected diversity reaches the output; (2) automated, general-purpose specification-level methods that inject diversity through intermediate generative configurations; and (3) empirical evidence across 5 open-ended tasks and 4 backbone models showing that specification-level injection improves output diversity over lower-level test-time baselines while maintaining comparable quality.

\section{Related Work}
\label{sec:related}

We discuss related work in three areas: test-time methods for diverse generation, planning-based generation, and diversity evaluation.

\paragraph{Decoding-based Diverse Generation.} The most straightforward approach to increasing output diversity is to modify the sampling strategy at decoding time. Temperature scaling flattens the output distribution to encourage more varied token selections 
\cite{ackley1985learning}. 
Top-$k$ sampling~\citep{fan2018hierarchical}, nucleus sampling~\citep{holtzman2020curious}, and min-$p$ sampling~\citep{minh2025minp} each restrict the candidate token pool using different truncation criteria. 
Contrastive search~\citep{su2023contrastive} penalizes tokens that are too similar to previous context, and diverse beam search~\citep{Vijayakumar2016DiverseBS} encourages diversity across beam groups. 
Avoidance Decoding~\citep{park-etal-2025-avoidance} extends this contrastive idea to multi-output generation by treating prior branches as negative samples and penalizing candidate tokens using hidden-state and sentence-embedding similarity. 
G2~\citep{ruan2025g2} takes a different approach by steering decoding with dual guide modules that encourage novelty and suppress repetition. These methods all operate at the token level and do not introduce explicit per-output conditioning that structures the diversity in generation.

\paragraph{Prompt-based Diverse Generation.}
Several training-free methods promote diversity through prompt design. 
\citet{zhang2024commonsense} improve diversity in commonsense generation tasks by diversifying in-context learning examples. 
\citet{hayati2024far} explore diverse perspective extraction through criteria-based prompting on subjective topics. 
Verbalized Sampling~\citep{zhang2025verbalizedsamplingmitigatemode} prompts the model to generate multiple candidate responses with associated probabilities in a single call. SSOT~\citep{misaki2026string} prepends a randomly generated string to each generation as a diversity seed, and \citet{agrawal2026addressing} prepend random, unrelated concepts to prompts and measure the resulting changes in responses. Unlike these methods, we make the diversity source explicit for each output before generation.

\paragraph{Planning-based Generation.} 
A separate line of work uses outlines or plans to control text structure, especially in storytelling, though not for the purpose of promoting output diversity. DOC~\citep{yang2023doc} uses detailed hierarchical outlines to improve coherence in long-form storytelling. \citet{wang2024guiding} combine symbolic outline planning with LLM-based surface realization for story generation, building on earlier narrative planning work~\citep{Riedl_2010,cardona2024story}. Luminate~\citep{suh2024luminate} generates stylistic and content dimensions for human-AI co-creative exploration, but requires interactive human steering. These methods usually target quality of the output rather than diversity and are mostly restricted to storytelling or require human involvement.

\paragraph{Diversity Evaluation} Measuring diversity in LLM outputs remains an open challenge. Standard diversity metrics include lexical measures such as Distinct-$N$~\citep{li2016diversity} and Self-BLEU~\citep{zhu2018texygen}, and embedding-based distances using pretrained sentence encoders~\citep{reimers2019sentence}. \citet{shypula2025evaluating} argue that diversity should be evaluated jointly with quality. Datasets have been developed recently to facilitate evaluation. \citet{zhang2025noveltybench} propose NoveltyBench, a benchmark specifically designed for evaluating response diversity. \citet{jiang2026artificial} introduce Infinite Chats, a large-scale dataset of open-ended queries and use it to study inter- and intra-model homogeneity in modern LLMs. 
We use these metrics and benchmarks in our experiments.

\section{Diversity Injection Framework}
\label{sec:framework}

When an LLM generates multiple outputs for the same prompt, the outputs may differ from one another. These differences can come from two sources: randomness in the decoding process, or explicit guidance that steers each generation in a different direction. We formalize this distinction into a unified framework.

\subsection{Formulation}
Given an input prompt $x$, a method produces $m$ outputs $\{y_1, \dots, y_m\}$. We write the generation of each output as:
\begin{equation}
    y_i = G(x, z_i)
\end{equation}
where $z_i$ is a \emph{diversity source}, the information beyond $x$ that the model conditions on when generating $y_i$. For some methods, $z_i$ is absent; for others, it may be a random string or a set of constraints. We categorize methods into three levels based on what information $z_i$ carries.

\paragraph{Level 0: No Injection.}
No per-output diversity source exists. Every output is sampled from the same conditional distribution: $y_i \sim P(Y| x)$, and differences between outputs arise entirely from the stochasticity of the decoding process.

\paragraph{Level 1: Surface-Level Injection.}
Each output is conditioned on a $z_i$ that carries no semantic content about the target output, such as a random seed string, a nonce token, or an arbitrary identifier. The diversity source perturbs the input but does not specify any property of the output.

\paragraph{Level 2: Specification-Level Injection.}
Each output is conditioned on a $z_i$ that specifies the generative configuration along multiple dimensions, such as narrative perspective, tone, focal subject, stylistic register, and content keywords. 
The diversity source establishes \emph{how} the model approaches generation and \emph{which elements} it foregrounds, jointly steering style and content.
 
The three levels form a spectrum of increasing output diversity in the conditioning signal. 
A method belongs to Level~1 if there exists a per-output $z_i$ that varies across outputs but carries no semantic content. It belongs to Level~2 if $z_i$ specifies meaningful properties of the output along multiple dimensions. 

\subsection{Classification of Existing Methods} \label{sec:classification}
Table~\ref{tab:method_classification} classifies existing diverse generation methods by injection level.

\begin{table}[t]
\centering
\caption{Classification of existing diverse output generation methods by injection level. }
\label{tab:method_classification}
\small
\setlength{\tabcolsep}{2.2pt} 
\begin{tabular}{llc}
\toprule
\textbf{Method} & \textbf{$z_i$} & \textbf{Level} \\
\midrule
Direct Generation & none & 0 \\

\shortstack[l]{Verbalized Sampling \\ \citep{zhang2025verbalizedsamplingmitigatemode}} & none & 0 \\
G2 \cite{ruan2025g2} & none & 0 \\

\midrule
SSOT \cite{misaki2026string} & random string & 1 \\
Rand Concept \cite{agrawal2026addressing} &random concept &1 \\

\bottomrule
\end{tabular}
\end{table}

\paragraph{Level~0 Methods.} 
\textit{Direct Generation} and its variants (temperature scaling, top-$k$ sampling, nucleus sampling, and min-$p$ sampling) repeatedly sample from the same prompt. They may reshape the token distribution but add no per-output conditioning, so diversity comes solely from sampling randomness.  
\textit{Verbalized Sampling} \citep{zhang2025verbalizedsamplingmitigatemode} elicits multiple candidates in a single call, but issues the same prompt every time, leaving decoding stochasticity as the sole source of variation. 
\textit{G2} \citep{ruan2025g2} steers decoding with diversity-promoting guide modules but likewise uses the same prompt for every output.
 
\paragraph{Level~1 Methods.}
\emph{SSOT} \citep{misaki2026string} instructs the model to first generate a random string, then condition on it as a seed to produce the output. The random string serves as a $z_i$ and differs across outputs. However, it carries no semantic content about the desired output's topic or organization. The same holds for Random Concept \cite{agrawal2026addressing}, which prepends an unrelated concept as the seed.  
 
\paragraph{The Level~2 Gap.}
To our knowledge, no existing test-time method provides per-output specifications to promote diversity in general open-ended generation. A related line of work on narrative planning conditions generation on outlines or plot structures \citep{yang2023doc, suh2024luminate}, but these methods use planning or specification mainly to improve the quality of a single generation and are often restricted to storytelling.
We address this gap in Section~\ref{sec:method} by introducing fully automated Level~2 methods that provide specifications to induce diverse outputs across general open-ended tasks.

\subsection{Quantifying Diversity Transmission}\label{sec:transmission}
 
 The framework above tells us \emph{where} diversity is injected. A natural next question is: how much of the injected diversity actually survives to the final output? A method might generate highly diverse plans (high diversity in $z$), but if the LLM ignores or flattens those plans during generation, the output diversity will be low. We now formalize a way to measure this.
 
Let $Z$ and $Y$ be random variables representing the diversity source and the output for a given prompt $X = x$. By the definition of conditional mutual information, the entropy of the output can be decomposed as:
\begin{equation}
    H(Y \mid X) = I(Y; Z \mid X) + H(Y \mid X, Z)
    \label{eq:decomposition}
\end{equation}
where $I(Y; Z \mid X)$ is the \textbf{realized diversity}, the portion of output diversity that is attributable to variation in $Z$. In other words, this is the diversity that the injection mechanism successfully transmits to the output. $H(Y \mid X, Z)$ is the \textbf{residual diversity}, the variation that arises even when the diversity source $z$ is held fixed, i.e., decoding randomness.
The realized diversity can be further expanded as:
\begin{equation}
    I(Y; Z \mid X) = H(Z \mid X) - H(Z \mid X, Y)
    \label{eq:realized}
\end{equation}
where $H(Z \mid X)$ is the \textbf{injected diversity}, measuring how diverse the sources themselves are, and $H(Z \mid X, Y)$ captures \textbf{information loss}, measuring the diversity present in $Z$ that is not realized in the output. Based on this decomposition, we define the transmission score as follows. 

\begin{definition}[Transmission Score] For methods with an explicit diversity source $Z$ and $H(Z\mid X)>0$, we define the transmission score as
    \begin{equation}
    T = \frac{I(Y; Z \mid X)}{H(Z \mid X)}
    \label{eq:transmission}
\end{equation}
It measures the fraction of source diversity that is realized in the final output.
\end{definition}

Since $I(Y;Z\mid X) \le H(Z\mid X)$, the transmission score lies in $[0,1]$. 
A value of $T=0$ indicates that the output is conditionally independent of the source given the prompt, while $T=1$ indicates that the source variation is fully recoverable from the output under the chosen representation.
 
The transmission score is defined only for methods with an explicit diversity source (Levels~1 and~2). For Level~0 methods, no per-output $z_i$ exists, so the decomposition does not apply, and output diversity arises from decoding randomness. We examine how $T$ and the decomposition terms vary across methods in Section~\ref{sec:experiments}. 

\subsection{Estimating Output Entropy and Transmission Score}
\label{sec:holdout}

Computing the transmission score $T$ and the decomposition in Eq.~\ref{eq:decomposition}-\ref{eq:realized} requires estimating $H(Y| X), H(Y| X, Z)$, and $H(Z |X)$. The latter two can be computed directly from the LLM's log probabilities given a fixed $z$ or a fixed prompt. The challenge is estimating $H(Y|X)$, the marginal output entropy, which requires marginalizing over $Z$.

\begin{algorithm}[t]
\caption{$\textsc{EstimateT}(X, P, M, L)$: Estimate the Transmission Score}
\label{alg:transmission-score}
\begin{algorithmic}[1]
\Require Task $X$, model $P$, estimation size $M$, evaluation size $L$
\Ensure Transmission score $\hat{T}$
\State Sample estimation set $\{Z_j\}_{j=1}^{M} \sim P(Z | X)$
\State Sample evaluation set $\{Z_\ell\}_{\ell=1}^{L} \sim P(Z | X)$
\State Generate $Y_\ell \sim P(Y \mid Z_\ell, X)$ for each $\ell$
\For{$\ell = 1, \ldots, L$}
    \State $\hat{P}(Y_\ell \mid X) \gets \frac{1}{M}\sum_{j=1}^{M} P(Y_\ell \mid Z_j, X)$
    \State $\hat{h}_\ell \gets -\frac{1}{|Y_\ell|} \log_2 \hat{P}(Y_\ell \mid X)$
\EndFor
\State $\hat{H}(Y \mid X) \gets \frac{1}{L}\sum_\ell \hat{h}_\ell$
\State $\hat{H}(Y| X, Z) \!\leftarrow\! \frac{1}{L}\sum_\ell \!\left[-\frac{1}{|Y_\ell|}\! \log_2\! P(Y_\ell| Z_\ell, X)\!\right]$
\State $\hat{H}(Z | X) \gets \frac{1}{L}\sum_\ell \left[-\frac{1}{|Z_\ell|} \log_2 P(Z_\ell | X)\right]$
\State \Return $\hat{T} = \frac{\hat{H}(Y \mid X) - \hat{H}(Y \mid X, Z)}{\hat{H}(Z \mid X)}$
\end{algorithmic}
\end{algorithm}

\begin{figure*}[t!] 
    \includegraphics[width=\linewidth]{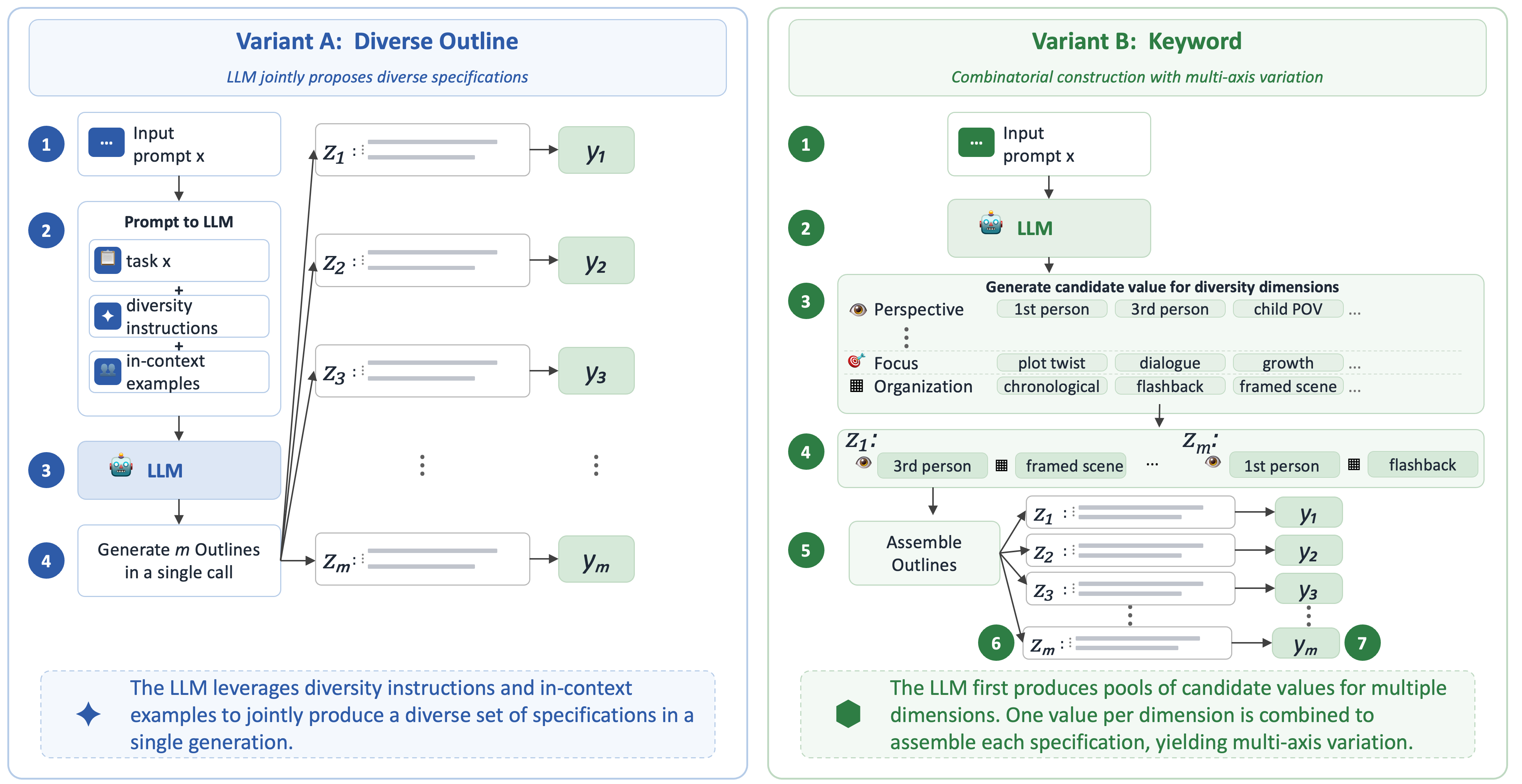}
    \caption{Two variants for generating diverse specifications ($z_i$'s) to encourage LLM outputs. Diverse Outline (left) relies on the LLM's internal capacity to jointly produce a batch of diverse outlines. Keyword (right) explicitly constructs combinatorial outlines by sampling keyword combinations from LLM generated outline contents.}
    \label{fig:method_plot}
\end{figure*}

We estimate $H(Y|X)$ using Monte Carlo approximation on a holdout set. For a given task $X$, we draw two independent sets of outlines from the LLM: an \textbf{estimation set} $\{Z_j\}_{j=1}^{M}$ and an \textbf{evaluation set} $\{Z_\ell\}_{\ell=1}^{L}$. For each $Z_\ell$, we generate one output $Y_\ell \sim P(Y \mid Z_\ell, X)$. We then approximate the marginal likelihood of each evaluation output by averaging over the estimation set:
$$\hat{P}(Y_\ell \mid X)= \frac{1}{M} \sum_{j=1}^{M} P(Y_\ell \mid Z_j, X).$$
The output entropy is then estimated by averaging the per-token log-probability across the $L$ evaluation members: $$\hat{H}(Y \mid X) = \frac{1}{L}\sum_{\ell=1}^{L} \left[-\frac{1}{|Y_\ell|} \log_2 \hat{P}(Y_\ell \mid X)\right],$$ 
where $|Y_\ell|$ is the number of tokens in $Y_\ell$. 

Note that $\hat{P}(Y_{\ell}|X)$ is an unbiased estimator. Since each $Z_j$ is drawn i.i.d.\ from $P(Z| X)$,
$\mathbb{E}_{\{Z_j\}}[\hat{P}(Y_\ell| X)] = \mathbb{E}_{Z \sim P(Z|X)}[P(Y_\ell | Z, X)] = P(Y_\ell | X).$
However, the entropy estimate $\hat{H}(Y \mid X)$ has a positive bias of $O(1/M)$ due to Jensen's inequality applied to the concave $\log$ function, so $\hat{H}(Y| X)$ is a soft upper bound on the true entropy. The bias vanishes as $M$ increases. Full derivations are provided in Appendix~\ref{app:bias_variance}.

Given estimates of $H(Y|X)$, $H(Y|X,Z)$, and $H(Z|X)$, we can calculate the transmission score. Algorithm~\ref{alg:transmission-score} summarizes the full procedure.

\section{Level-2 Instantiations}
\label{sec:method}

We propose two fully automated methods that instantiate Level~2 injection for general open-ended generation tasks. Both methods follow a two-step pipeline. Given an input prompt $x$, we first generate a set of per-output specifications $\{z_1,\ldots,z_m\}$, and then condition on each specification to produce a final output $y_i = G(x,z_i)$. Each specification is represented as a short outline describing the intended generative configuration, including dimensions such as perspective, tone, thematic focus, and content organization. The two methods differ in how these specifications are constructed.

\paragraph{Diverse Outline.}
Given task $x$, we prompt the LLM to generate $m$ specifications in a single call, where each specification is a compact list of 4-6 keywords capturing tone, format, perspective, and key focus. The prompt instructs the model to maximize diversity across specifications and specifies the output format (Figure~\ref{fig:method_plot} Left). This variant relies on the LLM's internal ability to propose diverse specifications. 

\paragraph{Keyword.} We prompt the LLM to generate a set of dimensions along which responses to the task can vary and a pool of candidate values for each (e.g., tone with values such as satirical or earnest). Each specification $z_i$ is assembled by selecting one value per dimension. To ensure the selected specifications differ from one another, we use greedy Hamming-distance maximization, iteratively selecting the combination most distant from all previously chosen ones (Figure~\ref{fig:method_plot} right).
This combinatorial construction encourages distinct outlines to differ across multiple axes.

Both methods require no human intervention and apply to open-ended generation task with the same pipeline. Prompts are provided in Appendix \ref{sec:appendix_prompt_and_result}.

\section{Experiments}
\label{sec:experiments}
Our evaluation addresses the following questions: (1) Does injecting more informative diversity sources lead to more diverse outputs, without sacrificing quality (Sec \ref{sec:main_results})?  (2) When a source is injected, does variation in the source contribute to output diversity, beyond what decoding randomness alone produces (Sec \ref{sec:Div_Decompose})?  (3) How does output diversity scale as we increase the number of generated outputs (Sec \ref{sec:ablations})?

We evaluate on five open-ended generation tasks spanning creative writing, conversation, poetry, argumentation, and general knowledge. 
\begin{itemize}[leftmargin=*]
    \item \textbf{WritingPrompts} \citep{huang2024gptwritingpromptsdatasetcomparativeanalysis} contains Reddit writing prompts that ask for short fiction given a premise,  spanning genres from science fiction to literary drama. 
    \item \textbf{Infinite Chats} \citep{jiang2026artificial} consists of 26K open-ended user queries covering diverse everyday topics, requiring substantive, multi-paragraph responses. 
    \item \textbf{NoveltyBench} \citep{zhang2025noveltybench} is a benchmark designed to evaluate response diversity, covering creative writing, factual knowledge, and subjective opinion generation with prompts that admit multiple valid answers. 
    \item \textbf{Poetry Foundation} \citep{poetry_foundation} provides poetry writing prompts with varied stylistic and thematic constraints. 
    \item \textbf{IBM ArgKP} \citep{bar-haim-etal-2020-arguments} contains debate topics for which the model must generate diverse argumentative key points. 
\end{itemize}
Unless otherwise noted, all experiments use the default decoding settings of each backbone model. We provide the generated outputs in Appendix \ref{sec:appendix_prompt_and_result}.

\begin{table*}[t]
\centering
\caption{Diversity and quality on Infinite Chats and NoveltyBench. Methods are grouped by injection level. G2 and High Temperature require access to the model and are evaluated only on Qwen3-4B. Bold indicates the best result. $\uparrow$ = higher is better. Level~2 methods achieve higher diversity while maintaining comparable quality.}
\label{tab:main_results}
\small
\begin{tabularx}{\textwidth}{l @{\hspace{4.5pt}} c @{\hspace{5.5pt}} l @{\hspace{8pt}} *{4}{>{\centering\arraybackslash}X} *{5}{>{\centering\arraybackslash}X}}
\toprule
 &  &  & \multicolumn{4}{c}{\textbf{Infinite Chats}} & \multicolumn{5}{c}{\textbf{NoveltyBench}} \\
\cmidrule(lr){4-7} \cmidrule(lr){8-12}
& \textbf{Lv.} & \textbf{Method} & LLM Div $\uparrow$ & Emb Div $\uparrow$ & Struct Div $\uparrow$ & Quality $\uparrow$ & LLM Div $\uparrow$ & Emb Div $\uparrow$ & Struct Div $\uparrow$ & Distinct-$N$ $\uparrow$ & Quality $\uparrow$\\
\midrule
\multirow{5}{*}{\rotatebox{90}{GPT-5-mini}} & 0 & Direct & 2.69{\tiny $\pm$1.52} & 0.26{\tiny $\pm$0.15} & 0.21{\tiny $\pm$0.09} & 8.30{\tiny $\pm$0.72} & 2.88{\tiny $\pm$1.29} & 0.41{\tiny $\pm$0.15} & 0.24{\tiny $\pm$0.09} & 15.25{\tiny $\pm$6.32} & 7.30{\tiny $\pm$0.57} \\
 & 0 & Verbalized & 2.84{\tiny $\pm$1.18} & 0.23{\tiny $\pm$0.11} & 0.20{\tiny $\pm$0.07} & \textbf{8.47{\tiny $\pm$0.45}} & 3.10{\tiny $\pm$1.21} & 0.36{\tiny $\pm$0.11} & 0.22{\tiny $\pm$0.07} & 9.76{\tiny $\pm$4.09} & 7.30{\tiny $\pm$0.62} \\
 & 1 & SSOT & 2.40{\tiny $\pm$0.78} & 0.18{\tiny $\pm$0.09} & 0.17{\tiny $\pm$0.04} & 8.30{\tiny $\pm$0.45} & 2.36{\tiny $\pm$0.81} & 0.24{\tiny $\pm$0.10} & 0.19{\tiny $\pm$0.06} & 4.74{\tiny $\pm$2.98} & 6.88{\tiny $\pm$0.74} \\
\cmidrule(lr){2-12}
 & \cellcolor{blue!10} 2 & \cellcolor{blue!10} Keyword (ours) & \cellcolor{blue!10} 4.87{\tiny $\pm$1.39} & \cellcolor{blue!10} 0.32{\tiny $\pm$0.13} & \cellcolor{blue!10} 0.25{\tiny $\pm$0.06} & \cellcolor{blue!10} 8.36{\tiny $\pm$0.54} & \cellcolor{blue!10} 6.63{\tiny $\pm$1.43} & \cellcolor{blue!10} 0.43{\tiny $\pm$0.09} & \cellcolor{blue!10} 0.31{\tiny $\pm$0.06} & \cellcolor{blue!10} 16.46{\tiny $\pm$3.78} & \cellcolor{blue!10} 7.32{\tiny $\pm$0.70} \\
 & \cellcolor{blue!10} 2 & \cellcolor{blue!10} Div Outline (ours) & \cellcolor{blue!10} \textbf{7.32{\tiny $\pm$1.55}} & \cellcolor{blue!10} \textbf{0.40{\tiny $\pm$0.13}} & \cellcolor{blue!10} \textbf{0.31{\tiny $\pm$0.08}} & \cellcolor{blue!10} 8.28{\tiny $\pm$0.61} & \cellcolor{blue!10} \textbf{7.13{\tiny $\pm$1.89}} & \cellcolor{blue!10} \textbf{0.50{\tiny $\pm$0.11}} & \cellcolor{blue!10} \textbf{0.35{\tiny $\pm$0.08}} & \cellcolor{blue!10} \textbf{16.75{\tiny $\pm$4.30}} & \cellcolor{blue!10} \textbf{7.39{\tiny $\pm$0.79}} \\
\midrule
\multirow{6}{*}{\rotatebox{90}{Qwen3-4B}} & 0 & Direct & 2.11{\tiny $\pm$0.85} & 0.19{\tiny $\pm$0.14} & 0.13{\tiny $\pm$0.07} & 7.02{\tiny $\pm$0.68} & 2.44{\tiny $\pm$0.80} & 0.30{\tiny $\pm$0.16} & 0.18{\tiny $\pm$0.10} & 10.96{\tiny $\pm$7.44} & 6.72{\tiny $\pm$0.81} \\
 & 0 & Verbalized & 2.31{\tiny $\pm$0.91} & 0.25{\tiny $\pm$0.15} & 0.13{\tiny $\pm$0.07} & 6.55{\tiny $\pm$1.04} & 2.77{\tiny $\pm$0.80} & 0.37{\tiny $\pm$0.14} & 0.18{\tiny $\pm$0.09} & 11.01{\tiny $\pm$5.09} & 6.26{\tiny $\pm$1.29} \\
 & 0 & Direct ($\tau{=}1.5$) & 2.35{\tiny $\pm$0.82} & 0.19{\tiny $\pm$0.14} & 0.13{\tiny $\pm$0.07} & 6.67{\tiny $\pm$0.99} & 2.36{\tiny $\pm$0.85} & 0.25{\tiny $\pm$0.15} & 0.17{\tiny $\pm$0.09} & 7.38{\tiny $\pm$5.77} & 6.51{\tiny $\pm$1.00} \\
  & 0 & G2 & 2.09{\tiny $\pm$0.69} & 0.12{\tiny $\pm$0.07} & 0.19{\tiny $\pm$0.05} & 7.03{\tiny $\pm$0.78} & 2.75{\tiny $\pm$0.62} & 0.21{\tiny $\pm$0.08} & 0.23{\tiny $\pm$0.06} & 5.54{\tiny $\pm$4.36} & 7.35{\tiny $\pm$0.74} \\

 & 1 & SSOT & 2.21{\tiny $\pm$0.66} & 0.15{\tiny $\pm$0.09} & 0.15{\tiny $\pm$0.06} & 6.70{\tiny $\pm$0.85} & 2.13{\tiny $\pm$0.74} & 0.20{\tiny $\pm$0.12} & 0.17{\tiny $\pm$0.08} & 6.03{\tiny $\pm$4.78} & 6.36{\tiny $\pm$1.05} \\
\cmidrule(lr){2-12}
 & \cellcolor{blue!10} 2 & \cellcolor{blue!10} Keyword (ours) & \cellcolor{blue!10} \textbf{5.77{\tiny $\pm$1.52}} & \cellcolor{blue!10} 0.33{\tiny $\pm$0.09} & \cellcolor{blue!10} 0.22{\tiny $\pm$0.06} & \cellcolor{blue!10} \textbf{7.71{\tiny $\pm$0.49}} & \cellcolor{blue!10} \textbf{5.01{\tiny $\pm$1.72}} & \cellcolor{blue!10} 0.38{\tiny $\pm$0.09} & \cellcolor{blue!10} 0.25{\tiny $\pm$0.08} & \cellcolor{blue!10} 13.21{\tiny $\pm$5.58} & \cellcolor{blue!10} \textbf{7.40{\tiny $\pm$1.20}} \\
 & \cellcolor{blue!10} 2 & \cellcolor{blue!10} Div Outline (ours) & \cellcolor{blue!10} 5.58{\tiny $\pm$1.82} & \cellcolor{blue!10} \textbf{0.35{\tiny $\pm$0.12}} & \cellcolor{blue!10} \textbf{0.24{\tiny $\pm$0.08}} & \cellcolor{blue!10} 7.45{\tiny $\pm$0.59} & \cellcolor{blue!10} 5.00{\tiny $\pm$2.00} & \cellcolor{blue!10} \textbf{0.41{\tiny $\pm$0.11}} & \cellcolor{blue!10} \textbf{0.27{\tiny $\pm$0.09}} & \cellcolor{blue!10} \textbf{13.85{\tiny $\pm$5.79}} & \cellcolor{blue!10} 7.09{\tiny $\pm$1.17} \\
\bottomrule
\end{tabularx}
\end{table*}

\subsection{Specification-Level Injection Improves Diversity with Comparable Quality}
\label{sec:main_results}

We compare six methods, including three Level~0 baselines (Direct Generation, Verbalized Sampling \cite{zhang2025verbalizedsamplingmitigatemode}, and G2 \cite{ruan2025g2}), one Level~1 method (SSOT \cite{misaki2026string}), and two Level~2 methods proposed by this work (Keyword and Diverse Outline). 

We use four backbone models: GPT-5-mini \cite{singh2026openaigpt5card}, Qwen3-4B \cite{yang2025qwen3technicalreport}, DeepSeek-V3.2 \cite{deepseekai2025deepseekv32pushingfrontieropen}, and Mistral-Large-3 \cite{mistralai_large3_2512}.  G2 requires access to model logits for its guide modules, which is only available for open-weight models. We therefore report G2 results with Qwen3-4B only. For open-weight models whose temperature can be set above one (Qwen3-4B and DeepSeek-V3.2), we additionally include a high-temperature baseline ($\tau = 1.5$). Detailed setup is elaborated in Appendix \ref{appendix:additional_exp_setting}.

We evaluate our experimental results along five dimensions. 
\begin{itemize}[leftmargin=*]
    \item \textbf{LLM-judged diversity}: for each pair of outputs from the same prompt, we ask GPT-5.3-Chat \citep{singh2026openaigpt5card} to rate their pairwise diversity on a 1--10 scale and average over all pairs.
    \item \textbf{Lexical diversity}: we report Distinct-3 (proportion of unique trigrams) and Self-BLEU (mean BLEU of each output against all others.
    \item \textbf{Embedding diversity}: we compute the mean pairwise cosine distance ($1 - \cos$) between sentence embeddings of all outputs from the same prompt, using OpenAI's text-embedding-3-small \citep{openai2024embeddings}. 
    \item \textbf{Structural diversity}: we first prompt GPT-5.3-Chat to extract a short outline from each output, then compute embedding diversity over the extracted outlines rather than the full outputs. This captures variation in high-level organization rather than surface wording. 
    \item \textbf{Quality}: we ask GPT-5.3-Chat to rate each output on a 1--10 scale for coherence, relevance, and overall quality, and report the average.
\end{itemize}
We sample 100 prompts per dataset for evaluation, except for ArgKP where we retain 41 prompts after filtering those containing harmful content. For each prompt, we generate 20 outputs and report the mean and standard deviation of each metric across prompts. For NoveltyBench, we additionally report its specialized Distinct-N metric, which counts the number of unique equivalence classes among generated responses.

Table~\ref{tab:main_results} shows that both Keyword and Diverse Outline consistently achieve the highest diversity across all metrics on both benchmarks, while maintaining comparable quality to baselines. This suggests that introducing more informative specifications is an effective way to improve diversity in open-ended generation. Additional results across diversity metrics, datasets, and backbone models are reported in Appendix~\ref{appendix:additional_experiment_full}.

\subsection{Varying Injected Source Expands Output Diversity}
\label{sec:Div_Decompose}
To understand why Level-2 methods outperform, we examine how variation in the injected source contributes to output diversity. We apply the entropy decomposition introduced in Sec \ref{sec:transmission} to measure the transmission score $T$ and realized diversity $I(Y;Z|X)$.  This analysis requires an explicit per-output diversity source $z_i$. Level~0 methods have no such source, and both terms are undefined for them. We therefore compare only methods with explicit $z_i$. Table~\ref{tab:T_values} reports these quantities alongside the output entropy $H(Y | X)$, estimated using the holdout procedure described in Sec \ref{sec:holdout}. 

As shown in Table~\ref{tab:T_values}, SSOT has near-zero transmission ($T \approx 0.003$) and near-zero realized diversity, with an output entropy $H(Y| X)$ of only 0.80, far below the Level~2 methods. Although SSOT introduces a per-output diversity source, its random strings carry little information about the desired output and are largely ignored during generation. In contrast, both Keyword and Diverse Outline achieve substantially higher transmission scores and realized diversity, with output entropy above 2.8. These results suggest that effective diversity injection requires both informative sources and reliable transmission from the source to the final output. More results are in Appendix \ref{appendix:additional_experiment_full}.

\begin{table}[t]
\centering
\caption{Transmission score $T$, realized diversity $I(Y;Z| X)$, and output entropy $H(Y| X)$ on Qwen3-4B. $T$ and $I(Y;Z|X)$ are undefined for Level~0 methods. Level~2 methods achieve higher $T$ and $I(Y;Z|X)$ than Level 1 method.}
\label{tab:T_values}
\small
\setlength{\tabcolsep}{2.5pt}  
\begin{tabular}{llccc}
\toprule
Data & Method & $T$ $\uparrow$ & $I(Y;Z|X)$ $\uparrow$ & $H(Y| X)$ $\uparrow$\\
\midrule
\multirow{3}{*}{\rotatebox{90}{\shortstack{Infinite\\Chats}}}
& \cellcolor{blue!10} Div Outline (ours) & \cellcolor{blue!10} 0.552 & \cellcolor{blue!10} 1.74 & \cellcolor{blue!10} 2.98 \\
& \cellcolor{blue!10} Keyword  (ours)  & \cellcolor{blue!10} 0.464          & \cellcolor{blue!10} 1.45 & \cellcolor{blue!10} 2.96 \\
& SSOT             & 0.003          & 0.00 & 0.80 \\
\midrule
\multirow{3}{*}{\rotatebox{90}{\shortstack{Novelty\\Bench}}}
& \cellcolor{blue!10} Div Outline (ours) & \cellcolor{blue!10} 0.507 & \cellcolor{blue!10} 1.43 & \cellcolor{blue!10} 2.88 \\
& \cellcolor{blue!10} Keyword  (ours)  & \cellcolor{blue!10} 0.476          & \cellcolor{blue!10} 1.36 & \cellcolor{blue!10} 2.92 \\
& SSOT             & 0.006          & 0.00 & 0.73 \\
\bottomrule
\end{tabular}
\end{table}

\begin{figure}[hpbt]
    \centering
    \includegraphics[width=\linewidth]{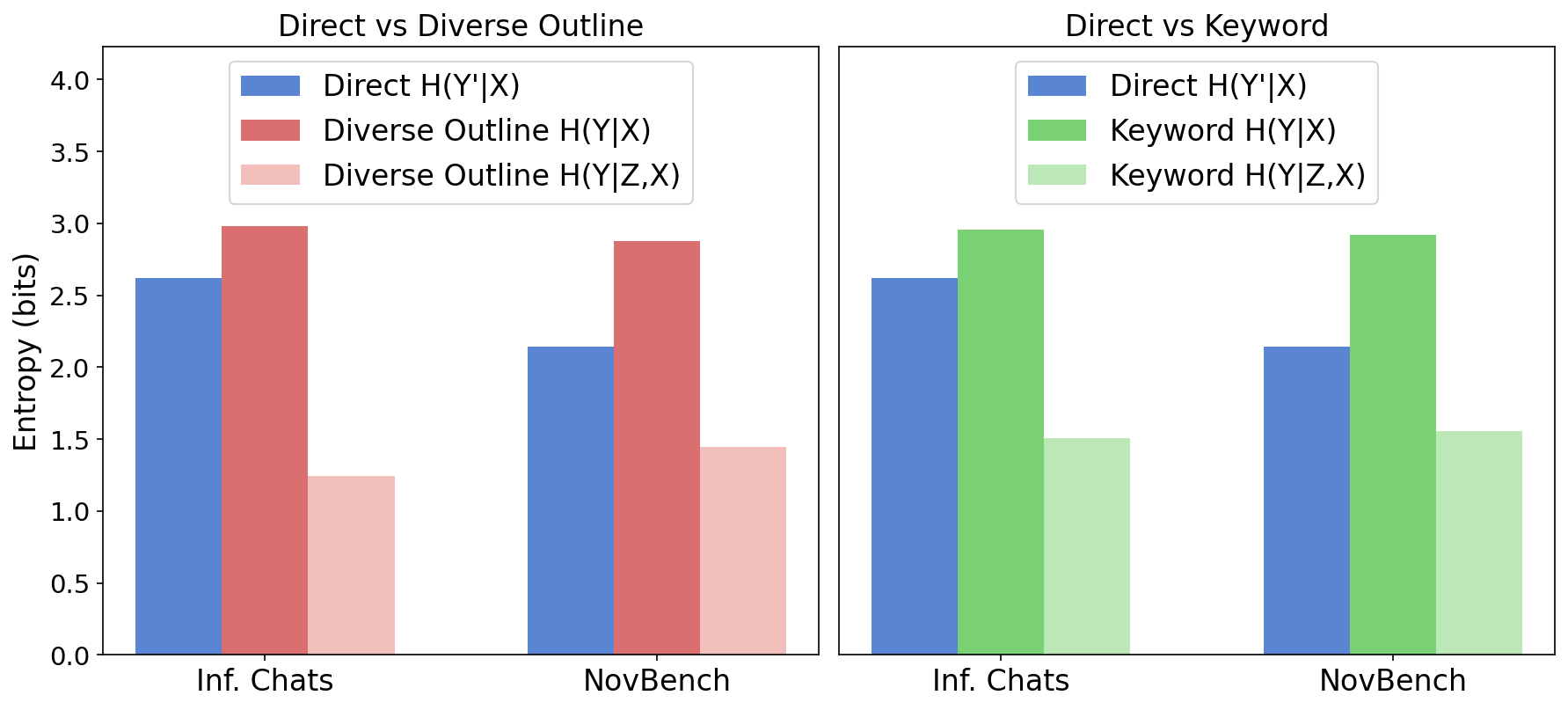}
    \caption{Entropy $H(Y'| X)$ of Direct Generation vs.\ outline-conditioned entropy $H(Y |X, Z)$ and total output entropy $H(Y|X)$ of Level 2 methods on NoveltyBench and Infinite Chats. Fixed-outline entropy falls below Direct Generation, yet total entropy exceeds it, showing that outline variation expands output diversity.}
    \label{fig:cond_entropy}
\end{figure}

Figure~\ref{fig:cond_entropy} compares the output entropy of Direct Generation $H(Y'|X)$ against the residual entropy $H(Y| X, Z)$ of our methods when conditioned on a single fixed outline, capturing per-outline decoding randomness, and the total output entropy of our methods when the outline varies $H(Y| X)$. 
For both Keyword and Diverse Outline, the total entropy $H(Y| X)$ exceeds the entropy of Direct Generation, while the fixed-source entropy $H(Y|X,Z)$ falls below it. This consistent ordering, $(H(Y|X,Z) < H(Y'| X) < H(Y|X)$, shows that conditioning on a single source narrows the per-source output distribution, yet varying the source across generations yields higher total diversity than Direct Generation. Thus, variation across injected sources contributes to the output diversity of specification-level methods.

We next provide a complementary analysis using embedding diversity. For each method, we vary the number of outlines $m \in \{2, 5, 10\}$, generating $k = 10, 4, 2$ outputs per outline respectively and measure two quantities: Embedding diversity for the same outline (\emph{within-outline diversity}), and Embedding diversity for different outlines (\emph{between-outline diversity}). If outline variation is the primary driver of diversity, between-outline diversity should exceed within-outline diversity.

\begin{figure}[hpbt]
    \centering
    \includegraphics[width=\linewidth]{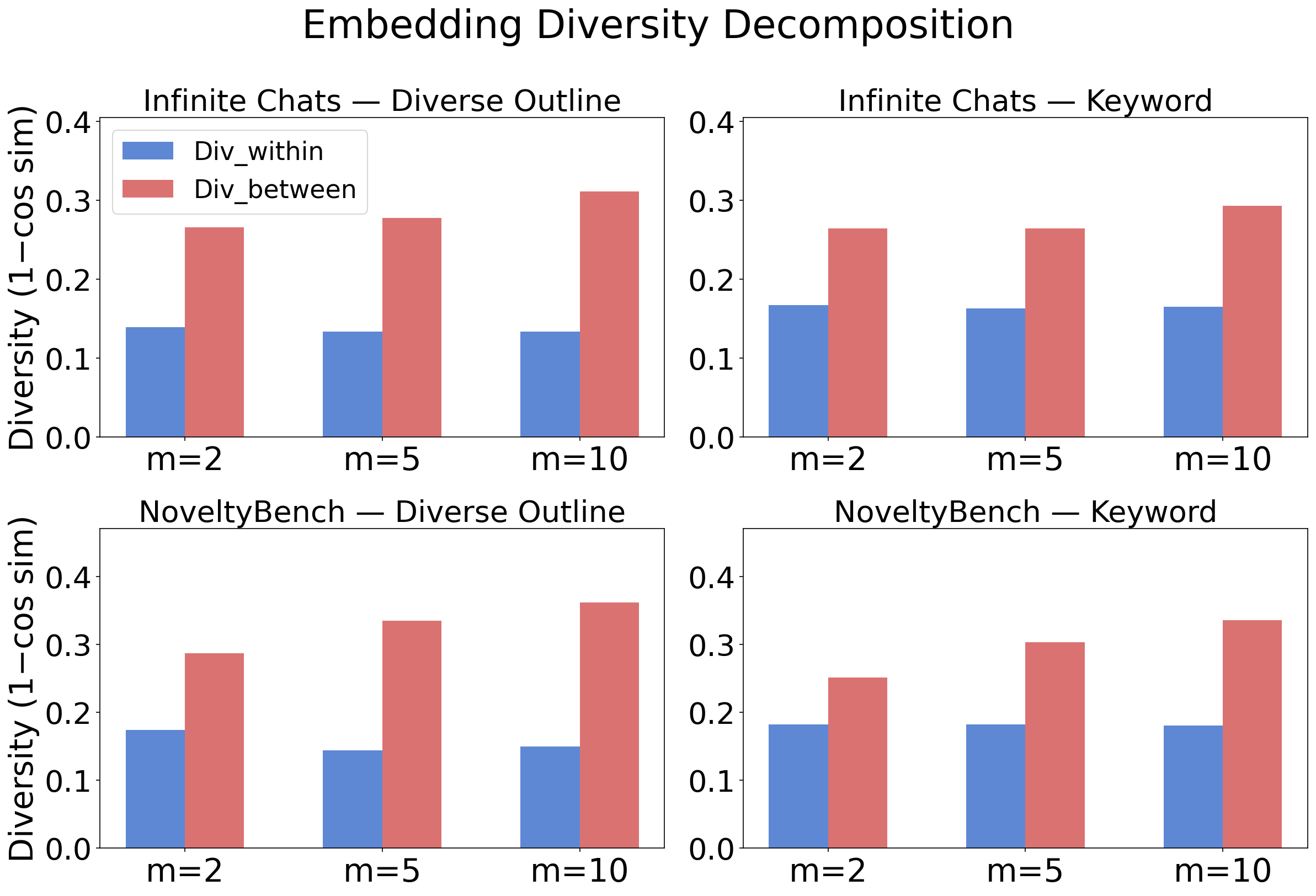}
    \caption{Within-outline diversity and between-outline diversity for Diverse Outline and Keyword, across $m \in \{2, 5, 10\}$ outlines on Infinite Chats and NoveltyBench. Between-outline diversity consistently exceeds within-outline diversity and grows with $m$, while within-outline diversity remains flat. More results are in Appendix \ref{appendix:additional_experiment_full}.}
    \label{fig:div_decomposition}
\end{figure}

Figure~\ref{fig:div_decomposition} shows that between-outline diversity consistently exceeds within-outline diversity across all values of $m$, for both methods and datasets. Within-outline diversity remains roughly stable as $m$ grows, indicating that repeated sampling under a fixed source contributes limited additional variation. Between-outline diversity increases with $m$, suggesting that adding more sources expands the range of generated outputs.

\begin{figure}[hpbt]
    \centering
    \includegraphics[width=\linewidth]{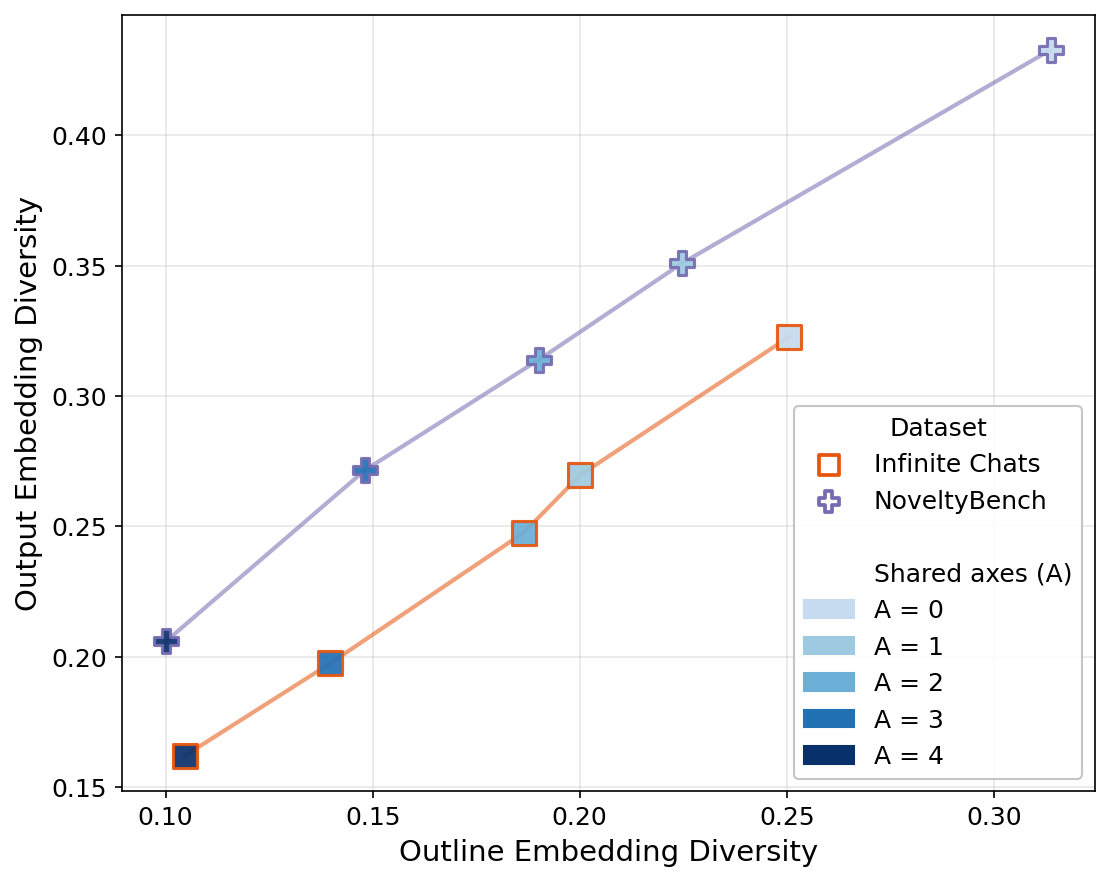}
\caption{Effect of Outline Diversity on Output Diversity. We reduce outline diversity in the Keyword method by forcing outlines to share values on an increasing number of dimensions ($0 \to 4$). Output embedding diversity scales linearly with specification embedding diversity on both NoveltyBench and Infinite Chats.} 
\label{fig:keyword_axis}
\end{figure}

To directly test whether outline diversity drives output diversity, we perform a controlled intervention on the keyword method. 
As described in Sec~\ref{sec:method}, Keyword constructs each specification by selecting one value per dimension. By default, each specification has independent values for all dimensions. We reduce specification diversity by forcing specifications to share the same value on a subset of dimensions. As more dimensions are shared, specifications become more similar. 
As shown in Figure\ref{fig:keyword_axis}, embedding diversity scales linearly with outline embedding diversity across both datasets, indicating that output variation is directly driven by the diversity of the injected specifications.

\subsection{Scaling with the Number of Generated Outputs}
\label{sec:ablations}

\begin{figure}[hpbt]
    \centering
    \includegraphics[width=\linewidth]{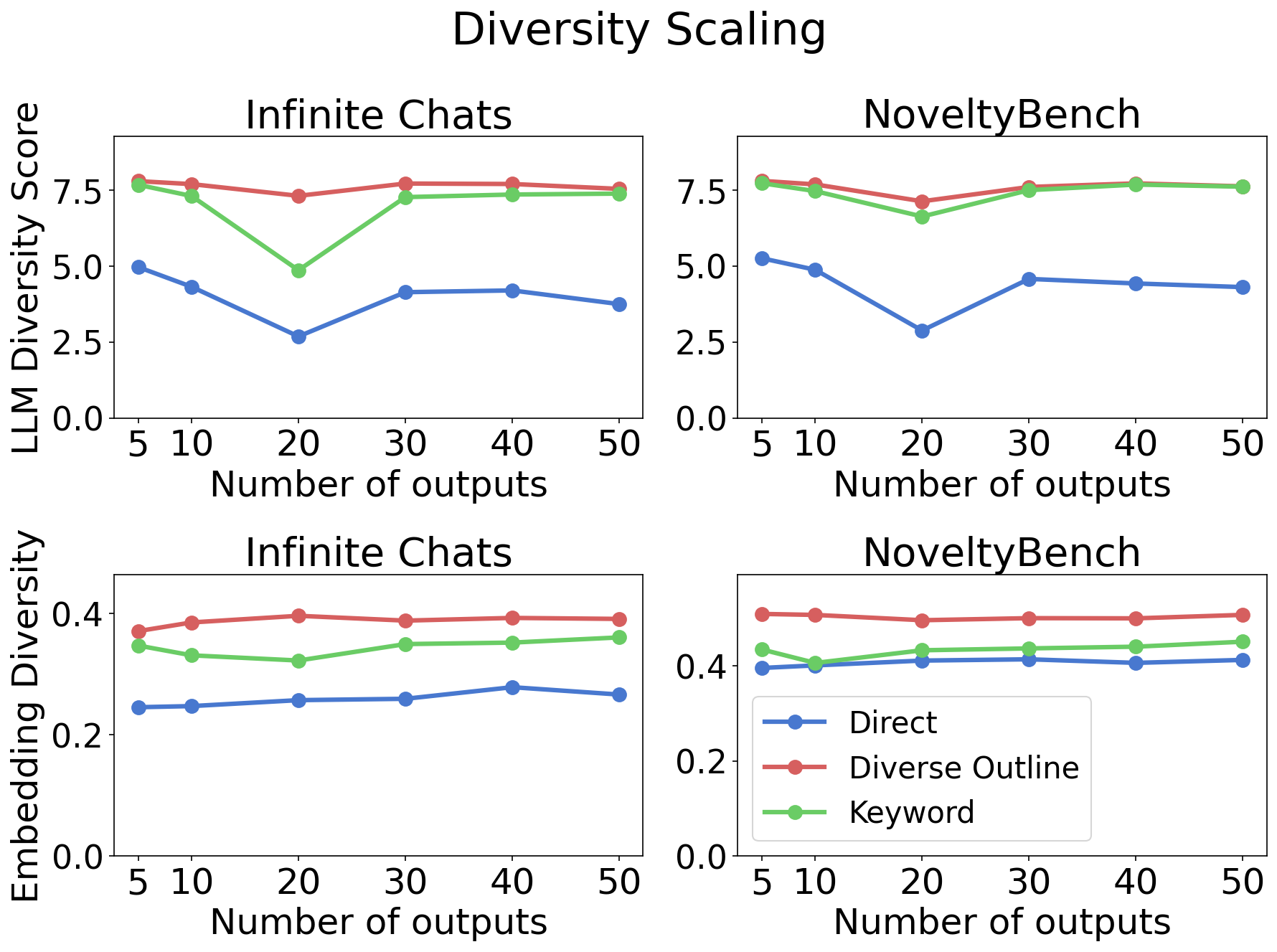}
    \caption{Diversity scaling as the number of generated outputs increases, using GPT-5-mini. Diverse Outline and Keyword maintain an advantage over Direct Generation across scales.}
    \label{fig:scaling_llm}
\end{figure}

We investigate how output diversity scales with the number of generated outputs. Using GPT-5-mini as the backbone, we vary the number of outputs across $\{5,10,20,30,40,50\}$ and compare Diverse Outline and Keyword against Direct Generation on Infinite Chats and NoveltyBench, reporting LLM-judged diversity and embedding diversity. 

As shown in Figure~\ref{fig:scaling_llm}, Diverse Outline and Keyword generally outperform Direct Generation across scales. In most cases, the gap remains visible as more outputs are generated. These results suggest that specification-level methods can make effective use of additional output slots by introducing diverse intermediate sources.

\section{Conclusion}
\label{sec:conclusion}
We introduced the Diversity Injection Framework, which unifies test-time diverse generation methods by the nature of their diversity source and quantifies diversity transmission via an entropy-based score. Guided by the framework, we proposed specification-level generation methods that inject diversity through detailed per-output configurations before generation begins. Diverse Outline lets the model propose all specifications in one pass and achieves high diversity in nearly all settings, while Keyword builds them from orthogonal axes, offering a more structured and interpretable construction. That both work through different source constructions shows that source design is the effective lever for diverse generation. Experiments across five open-ended tasks and four backbone models show that these methods substantially improve output diversity while maintaining comparable quality. We hope the framework and transmission score provide a foundation for diagnosing and improving diversity in future generation systems, and that specification-level injection offers a practical, training-free path toward richer, more varied LLM outputs.

\paragraph{Limitations.}
Our approach has the following limitations. First, specification-level injection adds a specification-generation step before output generation, which increases token usage and latency relative to Direct Generation, though in practice the step is lightweight and the cost amortizes when many outputs are requested. Second, our transmission score depends on a chosen representation of the diversity source and output. This is appropriate for comparing methods under a fixed evaluation protocol, but different representations may emphasize different aspects of source-to-output transmission. 

\bibliography{custom}

\newpage
\appendix
\FloatBarrier
\section{Additional Experiment Details}
\label{app:exp}
\subsection{Models and Evaluation Setup}
\label{appendix:additional_exp_setting}
We evaluate on four backbone models: \textbf{GPT-5-mini} (Azure API), \textbf{DeepSeek-V3.2} (Azure API), \textbf{Mistral-Large-3} (Azure API), and \textbf{Qwen3-4B} (served locally via vLLM).
All models use default decoding settings ($\tau=1.0$) unless stated otherwise.
For DeepSeek-V3.2 and Qwen3-4B-Instruct-2507, we additionally include a high-temperature baseline (Direct at $\tau{=}1.5$; Qwen additionally uses $\text{top-}p=0.9$).
Each experiment generates 20 outlines and one output per outline for  100 prompts per dataset. 

For the entropy estimation in Table \ref{tab:T_values} and Figure \ref{fig:cond_entropy}, we reuse the same 20 specifications and outputs in Table \ref{tab:main_results}. We randomly split the 20 specifications into an estimation set of 10 and an evaluation set of 10 to estimate $H(Y|X), H(Y|X, Z)$ and $H(Z|X)$ following the procedure in Section~\ref{sec:holdout}.

For the diversity decomposition experiment in Figure \ref{fig:div_decomposition}, we fix the total number of outputs at 20 and vary the number of outlines $m \in \{2, 5, 10\}$, generating 10, 4, and 2 outputs per outline, respectively.

LLM-based diversity and quality scoring is performed by \textbf{GPT-5.3-Chat} as the judge model.
Embedding diversity (both at the output level and the structural level) is computed using OpenAI's \textbf{text-embedding-3-small} model, measuring mean pairwise cosine distance $(1 - \text{cosine similarity})$ across the outputs.

All datasets and models used in this work are publicly available and used in accordance with their intended research use and terms of service. Any released code, prompts, and generated artifacts will be distributed for research purposes only.
\subsection{Additional Experiment Results}
\label{appendix:additional_experiment_full}

\paragraph{Additional Results for Sec \ref{sec:main_results}.} 
We provide additional results across datasets, backbone models, and metrics in this section.

Table~\ref{tab:main_results_remain} reports diversity and quality on NoveltyBench and Infinite Chat on Deepseek-V3.2 and Mistral-Large-3. The same pattern as in Table \ref{tab:main_results} holds. Level 2 methods achieve the highest diversity across most diversity metrics while maintaining comparable quality. 

Tables~\ref{tab:appendix_argkp_poetry} and~\ref{tab:appendix_wp} extend the evaluation to three more datasets: IBM ArgKP \cite{bar-haim-etal-2020-arguments}, Poetry Foundation \cite{poetry_foundation}, and WritingPrompts \cite{huang2024gptwritingpromptsdatasetcomparativeanalysis}.
The same evaluation protocol is applied.
Level 2 methods consistently outperform Level 0 and Level 1 baselines on all three diversity metrics (LLM Div, Emb Div, Struct Div) across models and datasets, while maintaining competitive quality scores.

Table~\ref{tab:lexical_div_main} and Table~\ref{tab:lexical_div_appendix} report lexical diversity, including Distinct-3, which measures the proportion of unique trigrams across outputs, and Self-Bleu for all methods across all datasets and models. Level-2 methods generally improve or match lexical diversity across most settings. For DeepSeek-V3.2, high temperature produces high token-level randomness, resulting in lower Self-BLEU. However, this comes without the structural diversity gains that Level-2 methods provide (Tables \ref{tab:main_results_remain}-\ref{tab:appendix_wp}).

\paragraph{Additional Results for Sec \ref{sec:Div_Decompose}.}
Table~\ref{tab:T_values_appendix} extends the transmission-score analysis to WritingPrompts, Poetry Foundation, and IBM ArgKP, reporting $T$, $I(Y;Z\mid X)$, and $H(Y\mid X)$ for Diverse Outline, Keyword, and SSOT. Level-2 methods achieve substantially higher transmission scores and realized diversity than SSOT across all three datasets.

\begin{figure}[htbp]
    \centering
    \includegraphics[width=\linewidth]{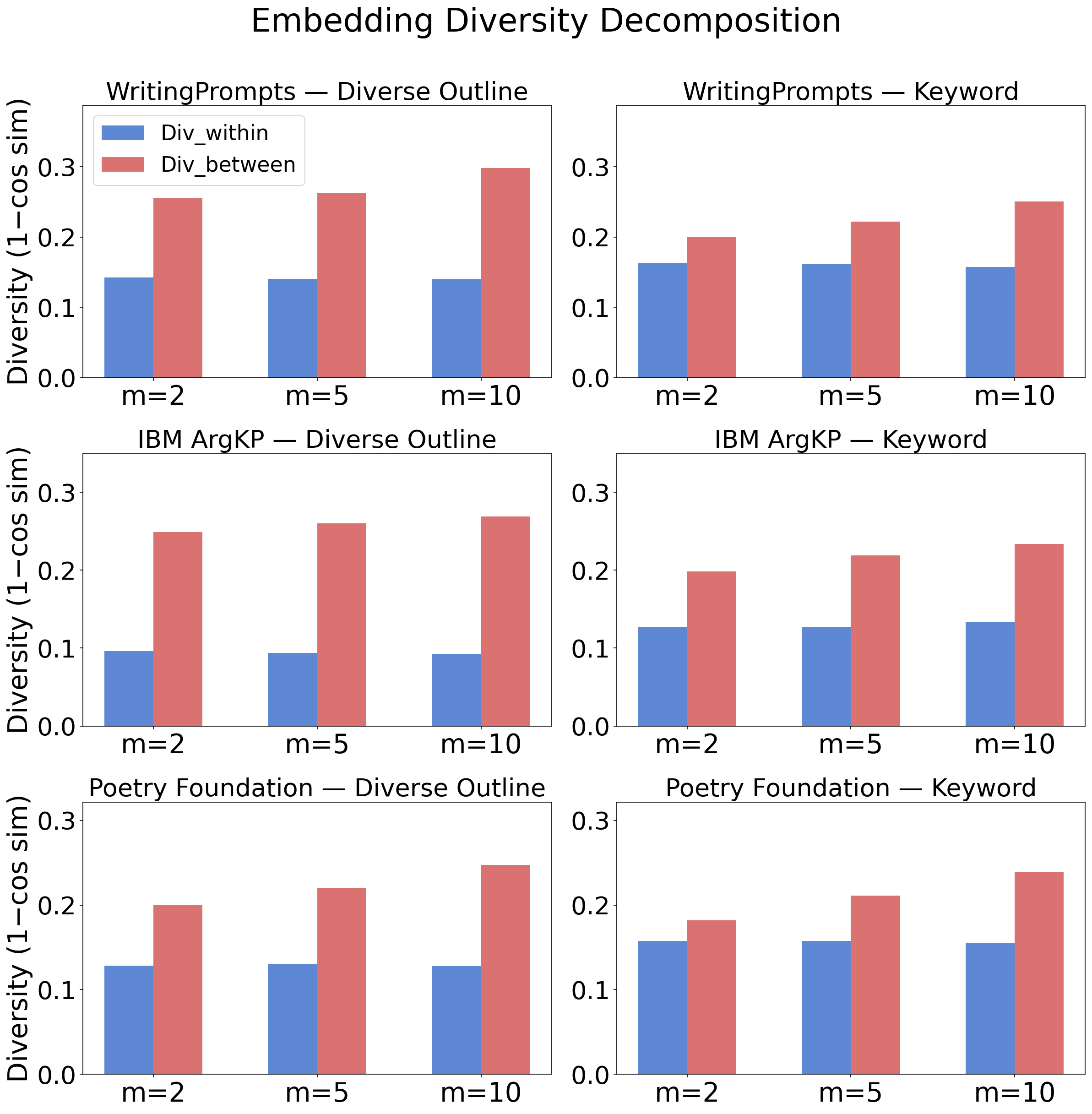}
        \caption{Within-outline diversity and between-outline diversity for Diverse Outline and Keyword, across $m \in \{2, 5, 10\}$ outlines on WritingPrompts, ArgKP and Poetry Foundation. Between-outline diversity consistently exceeds within-outline diversity and grows with $m$, while within-outline diversity remains flat. }
    \label{fig:div_decomposition_appendix}
\end{figure}

Additionally, Figure~\ref{fig:div_decomposition_appendix} shows the within-outline and between-outline embedding diversity decomposition for Diverse Outline and Keyword on these datasets. The same pattern as in Figure \ref{fig:div_decomposition} holds: between-outline diversity consistently exceeds within-outline diversity and grows with the number of outlines, while within-outline diversity remains flat, indicating that source variation contributes to the diversity gains of Level 2 methods across task types.

\begin{table*}[htbp]
\centering
\caption{Diversity and quality on Infinite Chats and NoveltyBench using DeepSeek-V3.2 and Mistral-Large-3. Methods are grouped by injection level. Bold indicates the best result. $\uparrow$ = higher is better.}
\label{tab:main_results_remain}
\small
\begin{tabularx}{\textwidth}{l @{\hspace{4.5pt}} c @{\hspace{5.5pt}} l @{\hspace{8pt}} *{4}{>{\centering\arraybackslash}X} *{5}{>{\centering\arraybackslash}X}}
\toprule
 &  &  & \multicolumn{4}{c}{\textbf{Infinite Chats}} & \multicolumn{5}{c}{\textbf{NoveltyBench}} \\
\cmidrule(lr){4-7} \cmidrule(lr){8-12}
& \textbf{Lv.} & \textbf{Method} & LLM Div $\uparrow$ & Emb Div $\uparrow$ & Struct Div $\uparrow$ & Quality $\uparrow$ & LLM Div $\uparrow$ & Emb Div $\uparrow$ & Struct Div $\uparrow$ & Distinct-$N$ $\uparrow$ & Quality $\uparrow$\\
\midrule
\multirow{6}{*}{\rotatebox{90}{DeepSeek-V3.2}} & 0 & Direct & 3.37{\tiny $\pm$1.59} & 0.29{\tiny $\pm$0.16} & 0.21{\tiny $\pm$0.07} & \textbf{7.51{\tiny $\pm$0.55}} & 3.39{\tiny $\pm$1.44} & 0.42{\tiny $\pm$0.16} & 0.25{\tiny $\pm$0.08} & 17.87{\tiny $\pm$4.00} & 7.39{\tiny $\pm$0.49} \\
 & 0 & Direct ($\tau{=}1.5$) & 3.36{\tiny $\pm$1.65} & 0.32{\tiny $\pm$0.13} & 0.21{\tiny $\pm$0.08} & 7.07{\tiny $\pm$0.73} & 3.59{\tiny $\pm$1.51} & 0.43{\tiny $\pm$0.14} & 0.25{\tiny $\pm$0.09} & \textbf{19.28{\tiny $\pm$2.64}} & 6.90{\tiny $\pm$0.82} \\
 & 0 & Verbalized & 3.03{\tiny $\pm$1.30} & 0.31{\tiny $\pm$0.15} & 0.19{\tiny $\pm$0.06} & 6.88{\tiny $\pm$1.00} & 3.22{\tiny $\pm$1.14} & 0.39{\tiny $\pm$0.15} & 0.22{\tiny $\pm$0.06} & 11.55{\tiny $\pm$4.11} & 6.79{\tiny $\pm$0.72} \\
 & 1 & SSOT & 2.86{\tiny $\pm$0.96} & 0.26{\tiny $\pm$0.16} & 0.19{\tiny $\pm$0.06} & 6.97{\tiny $\pm$1.09} & 3.25{\tiny $\pm$0.77} & 0.38{\tiny $\pm$0.13} & 0.22{\tiny $\pm$0.07} & 11.18{\tiny $\pm$4.40} & 6.53{\tiny $\pm$1.28} \\
\cmidrule(lr){2-12}
 & \cellcolor{blue!10} 2 & \cellcolor{blue!10} Keyword (ours) & \cellcolor{blue!10} 5.88{\tiny $\pm$1.44} & \cellcolor{blue!10} 0.40{\tiny $\pm$0.13} & \cellcolor{blue!10} 0.25{\tiny $\pm$0.06} & \cellcolor{blue!10} 7.50{\tiny $\pm$0.81} & \cellcolor{blue!10} 6.77{\tiny $\pm$1.85} & \cellcolor{blue!10} 0.47{\tiny $\pm$0.10} & \cellcolor{blue!10} 0.29{\tiny $\pm$0.07} & \cellcolor{blue!10} 17.41{\tiny $\pm$3.75} & \cellcolor{blue!10} \textbf{7.77{\tiny $\pm$0.80}} \\
 & \cellcolor{blue!10} 2 & \cellcolor{blue!10} Div Outline (ours) & \cellcolor{blue!10} \textbf{7.07{\tiny $\pm$1.76}} & \cellcolor{blue!10} \textbf{0.44{\tiny $\pm$0.13}} & \cellcolor{blue!10} \textbf{0.27{\tiny $\pm$0.07}} & \cellcolor{blue!10} 7.36{\tiny $\pm$0.88} & \cellcolor{blue!10} \textbf{7.13{\tiny $\pm$2.12}} & \cellcolor{blue!10} \textbf{0.50{\tiny $\pm$0.11}} & \cellcolor{blue!10} \textbf{0.32{\tiny $\pm$0.08}} & \cellcolor{blue!10} 18.04{\tiny $\pm$3.33} & \cellcolor{blue!10} 7.43{\tiny $\pm$0.97} \\
\midrule
\multirow{5}{*}{\rotatebox{90}{Mistral-L3}} & 0 & Direct & 2.33{\tiny $\pm$0.88} & 0.28{\tiny $\pm$0.17} & 0.13{\tiny $\pm$0.06} & 6.92{\tiny $\pm$0.68} & 2.59{\tiny $\pm$0.87} & 0.45{\tiny $\pm$0.13} & 0.16{\tiny $\pm$0.07} & 14.48{\tiny $\pm$5.77} & 6.80{\tiny $\pm$0.58} \\
 & 0 & Verbalized & 2.63{\tiny $\pm$0.94} & 0.26{\tiny $\pm$0.15} & 0.12{\tiny $\pm$0.05} & 6.91{\tiny $\pm$0.85} & 3.12{\tiny $\pm$1.02} & 0.37{\tiny $\pm$0.13} & 0.15{\tiny $\pm$0.07} & 9.69{\tiny $\pm$4.17} & 6.81{\tiny $\pm$0.62} \\
 & 1 & SSOT & 2.78{\tiny $\pm$0.72} & 0.21{\tiny $\pm$0.12} & 0.12{\tiny $\pm$0.05} & 7.26{\tiny $\pm$0.73} & 2.70{\tiny $\pm$0.76} & 0.30{\tiny $\pm$0.12} & 0.14{\tiny $\pm$0.06} & 9.77{\tiny $\pm$4.28} & 6.74{\tiny $\pm$1.07} \\
\cmidrule(lr){2-12}
 & \cellcolor{blue!10} 2 & \cellcolor{blue!10} Keyword (ours) & \cellcolor{blue!10} 6.51{\tiny $\pm$1.32} & \cellcolor{blue!10} 0.35{\tiny $\pm$0.11} & \cellcolor{blue!10} 0.19{\tiny $\pm$0.05} & \cellcolor{blue!10} \textbf{8.02{\tiny $\pm$0.40}} & \cellcolor{blue!10} 7.33{\tiny $\pm$1.18} & \cellcolor{blue!10} 0.42{\tiny $\pm$0.08} & \cellcolor{blue!10} 0.23{\tiny $\pm$0.06} & \cellcolor{blue!10} 15.13{\tiny $\pm$5.62} & \cellcolor{blue!10} \textbf{8.16{\tiny $\pm$0.52}} \\
 & \cellcolor{blue!10} 2 & \cellcolor{blue!10} Div Outline (ours) & \cellcolor{blue!10} \textbf{7.65{\tiny $\pm$1.67}} & \cellcolor{blue!10} \textbf{0.40{\tiny $\pm$0.12}} & \cellcolor{blue!10} \textbf{0.23{\tiny $\pm$0.07}} & \cellcolor{blue!10} 7.86{\tiny $\pm$0.46} & \cellcolor{blue!10} \textbf{7.84{\tiny $\pm$1.72}} & \cellcolor{blue!10} \textbf{0.46{\tiny $\pm$0.10}} & \cellcolor{blue!10} \textbf{0.26{\tiny $\pm$0.08}} & \cellcolor{blue!10} \textbf{15.74{\tiny $\pm$5.46}} & \cellcolor{blue!10} 7.91{\tiny $\pm$0.89} \\
\bottomrule
\end{tabularx}
\end{table*}

\begin{table*}[htbp]
\centering
\caption{Diversity and quality on IBM ArgKP and Poetry Foundation. Methods are grouped by injection level. DeepSeek-V3.2 and Qwen3-4B include a high-temperature baseline (Direct at $\tau{=}1.5$). 
\textbf{Bold} = best per column within each model. $\uparrow$ = higher is better.}
\label{tab:appendix_argkp_poetry}
\small
\begin{tabularx}{\textwidth}{lll *{4}{>{\centering\arraybackslash}X} *{4}{>{\centering\arraybackslash}X}}
\toprule
 &  &  & \multicolumn{4}{c}{\textbf{IBM ArgKP}} & \multicolumn{4}{c}{\textbf{Poetry Foundation}} \\
\cmidrule(lr){4-7} \cmidrule(lr){8-11}
& \textbf{Lv.} & \textbf{Method} & LLM Div $\uparrow$ & Emb Div $\uparrow$ & Struct Div $\uparrow$ & Quality $\uparrow$ & LLM Div $\uparrow$ & Emb Div $\uparrow$ & Struct Div $\uparrow$ & Quality $\uparrow$\\
\midrule
\multirow{5}{*}{\rotatebox{90}{GPT-5-mini}} & 0 & Direct & 3.51{\tiny $\pm$1.09} & 0.23{\tiny $\pm$0.06} & 0.23{\tiny $\pm$0.04} & 7.26{\tiny $\pm$0.41} & 2.44{\tiny $\pm$0.85} & 0.11{\tiny $\pm$0.04} & 0.13{\tiny $\pm$0.04} & 8.30{\tiny $\pm$0.25} \\
 & 0 & Verbalized & 2.95{\tiny $\pm$0.76} & 0.19{\tiny $\pm$0.05} & 0.22{\tiny $\pm$0.05} & 7.30{\tiny $\pm$0.43} & 2.99{\tiny $\pm$0.89} & 0.11{\tiny $\pm$0.04} & 0.15{\tiny $\pm$0.03} & \textbf{8.40{\tiny $\pm$0.25}} \\
 & 1 & SSOT & 2.51{\tiny $\pm$0.47} & 0.15{\tiny $\pm$0.07} & 0.18{\tiny $\pm$0.04} & 6.93{\tiny $\pm$0.54} & 3.15{\tiny $\pm$0.67} & 0.10{\tiny $\pm$0.02} & 0.13{\tiny $\pm$0.03} & 8.32{\tiny $\pm$0.21} \\
\cmidrule(lr){2-11}
 & \cellcolor{blue!10} 2 & \cellcolor{blue!10} Keyword (ours) & \cellcolor{blue!10} 6.17{\tiny $\pm$0.92} & \cellcolor{blue!10} 0.28{\tiny $\pm$0.04} & \cellcolor{blue!10} 0.31{\tiny $\pm$0.05} & \cellcolor{blue!10} 7.32{\tiny $\pm$0.20} & \cellcolor{blue!10} 7.98{\tiny $\pm$1.25} & \cellcolor{blue!10} 0.22{\tiny $\pm$0.06} & \cellcolor{blue!10} 0.24{\tiny $\pm$0.04} & \cellcolor{blue!10} 7.94{\tiny $\pm$0.85} \\
 & \cellcolor{blue!10} 2 & \cellcolor{blue!10} Div Outline (ours) & \cellcolor{blue!10} \textbf{7.97{\tiny $\pm$1.07}} & \cellcolor{blue!10} \textbf{0.33{\tiny $\pm$0.07}} & \cellcolor{blue!10} \textbf{0.34{\tiny $\pm$0.05}} & \cellcolor{blue!10} \textbf{7.37{\tiny $\pm$0.30}} & \cellcolor{blue!10} \textbf{8.83{\tiny $\pm$1.29}} & \cellcolor{blue!10} \textbf{0.23{\tiny $\pm$0.06}} & \cellcolor{blue!10} \textbf{0.27{\tiny $\pm$0.04}} & \cellcolor{blue!10} 8.18{\tiny $\pm$0.85} \\
\midrule
\multirow{6}{*}{\rotatebox{90}{DeepSeek-V3.2}} & 0 & Direct & 3.71{\tiny $\pm$1.32} & 0.30{\tiny $\pm$0.05} & 0.21{\tiny $\pm$0.05} & 7.44{\tiny $\pm$0.21} & 2.90{\tiny $\pm$1.26} & 0.23{\tiny $\pm$0.08} & 0.17{\tiny $\pm$0.04} & 7.99{\tiny $\pm$0.32} \\
 & 0 & Direct ($\tau{=}1.5$) & 3.58{\tiny $\pm$1.27} & 0.30{\tiny $\pm$0.08} & 0.23{\tiny $\pm$0.05} & 7.26{\tiny $\pm$0.33} & 2.80{\tiny $\pm$1.02} & 0.25{\tiny $\pm$0.08} & 0.15{\tiny $\pm$0.04} & 7.73{\tiny $\pm$0.37} \\
 & 0 & Verbalized & 2.32{\tiny $\pm$0.48} & 0.20{\tiny $\pm$0.03} & 0.18{\tiny $\pm$0.04} & 7.36{\tiny $\pm$0.25} & 3.23{\tiny $\pm$0.95} & 0.25{\tiny $\pm$0.07} & 0.16{\tiny $\pm$0.04} & 8.01{\tiny $\pm$0.27} \\
 & 1 & SSOT & 2.29{\tiny $\pm$0.54} & 0.13{\tiny $\pm$0.02} & 0.14{\tiny $\pm$0.03} & 7.53{\tiny $\pm$0.25} & 4.26{\tiny $\pm$1.17} & 0.16{\tiny $\pm$0.05} & 0.17{\tiny $\pm$0.04} & 7.97{\tiny $\pm$0.36} \\
\cmidrule(lr){2-11}
 & \cellcolor{blue!10} 2 & \cellcolor{blue!10} Keyword (ours) & \cellcolor{blue!10} 5.32{\tiny $\pm$1.30} & \cellcolor{blue!10} 0.29{\tiny $\pm$0.05} & \cellcolor{blue!10} 0.24{\tiny $\pm$0.04} & \cellcolor{blue!10} \textbf{7.73{\tiny $\pm$0.23}} & \cellcolor{blue!10} 7.39{\tiny $\pm$1.44} & \cellcolor{blue!10} 0.30{\tiny $\pm$0.06} & \cellcolor{blue!10} 0.22{\tiny $\pm$0.05} & \cellcolor{blue!10} 8.10{\tiny $\pm$0.25} \\
 & \cellcolor{blue!10} 2 & \cellcolor{blue!10} Div Outline (ours) & \cellcolor{blue!10} \textbf{7.66{\tiny $\pm$1.52}} & \cellcolor{blue!10} \textbf{0.35{\tiny $\pm$0.05}} & \cellcolor{blue!10} \textbf{0.30{\tiny $\pm$0.05}} & \cellcolor{blue!10} 7.71{\tiny $\pm$0.19} & \cellcolor{blue!10} \textbf{8.78{\tiny $\pm$1.03}} & \cellcolor{blue!10} \textbf{0.37{\tiny $\pm$0.07}} & \cellcolor{blue!10} \textbf{0.23{\tiny $\pm$0.05}} & \cellcolor{blue!10} \textbf{8.18{\tiny $\pm$0.19}} \\
\midrule
\multirow{5}{*}{\rotatebox{90}{Mistral-L3}} & 0 & Direct & 2.60{\tiny $\pm$0.93} & 0.24{\tiny $\pm$0.09} & 0.11{\tiny $\pm$0.06} & 6.66{\tiny $\pm$0.37} & 1.92{\tiny $\pm$0.38} & 0.14{\tiny $\pm$0.05} & 0.08{\tiny $\pm$0.03} & 7.61{\tiny $\pm$0.34} \\
 & 0 & Verbalized & 2.22{\tiny $\pm$0.43} & 0.14{\tiny $\pm$0.03} & 0.10{\tiny $\pm$0.04} & 7.14{\tiny $\pm$0.27} & 2.61{\tiny $\pm$0.60} & 0.19{\tiny $\pm$0.05} & 0.10{\tiny $\pm$0.03} & 7.76{\tiny $\pm$0.26} \\
 & 1 & SSOT & 2.21{\tiny $\pm$0.37} & 0.09{\tiny $\pm$0.02} & 0.09{\tiny $\pm$0.02} & 7.39{\tiny $\pm$0.25} & 4.17{\tiny $\pm$0.94} & 0.12{\tiny $\pm$0.03} & 0.10{\tiny $\pm$0.03} & 8.01{\tiny $\pm$0.36} \\
\cmidrule(lr){2-11}
 & \cellcolor{blue!10} 2 & \cellcolor{blue!10} Keyword (ours) & \cellcolor{blue!10} 5.79{\tiny $\pm$1.51} & \cellcolor{blue!10} 0.26{\tiny $\pm$0.04} & \cellcolor{blue!10} 0.17{\tiny $\pm$0.03} & \cellcolor{blue!10} \textbf{7.94{\tiny $\pm$0.22}} & \cellcolor{blue!10} 6.96{\tiny $\pm$0.98} & \cellcolor{blue!10} 0.20{\tiny $\pm$0.05} & \cellcolor{blue!10} 0.15{\tiny $\pm$0.04} & \cellcolor{blue!10} 8.18{\tiny $\pm$0.31} \\
 & \cellcolor{blue!10} 2 & \cellcolor{blue!10} Div Outline (ours) & \cellcolor{blue!10} \textbf{7.93{\tiny $\pm$1.90}} & \cellcolor{blue!10} \textbf{0.37{\tiny $\pm$0.05}} & \cellcolor{blue!10} \textbf{0.21{\tiny $\pm$0.03}} & \cellcolor{blue!10} 7.81{\tiny $\pm$0.28} & \cellcolor{blue!10} \textbf{8.03{\tiny $\pm$1.17}} & \cellcolor{blue!10} \textbf{0.26{\tiny $\pm$0.06}} & \cellcolor{blue!10} \textbf{0.16{\tiny $\pm$0.04}} & \cellcolor{blue!10} \textbf{8.19{\tiny $\pm$0.28}} \\
\midrule
\multirow{7}{*}{\rotatebox{90}{Qwen3-4B}} & 0 & Direct & 1.72{\tiny $\pm$0.40} & 0.14{\tiny $\pm$0.04} & 0.12{\tiny $\pm$0.03} & 6.89{\tiny $\pm$0.33} & 2.46{\tiny $\pm$0.52} & 0.10{\tiny $\pm$0.06} & 0.11{\tiny $\pm$0.03} & 7.47{\tiny $\pm$0.77} \\
 & 0 & Direct ($\tau{=}1.5$) & 2.02{\tiny $\pm$0.82} & 0.14{\tiny $\pm$0.07} & 0.12{\tiny $\pm$0.03} & 6.78{\tiny $\pm$0.52} & 2.83{\tiny $\pm$0.57} & 0.10{\tiny $\pm$0.05} & 0.12{\tiny $\pm$0.04} & 7.13{\tiny $\pm$0.87} \\
 & 0 & G2 & 1.59{\tiny $\pm$0.41} & 0.07{\tiny $\pm$0.02}& 0.16{\tiny $\pm$0.05} & 7.03{\tiny $\pm$0.25} & 2.46{\tiny $\pm$0.48} & 0.08{\tiny $\pm$0.03} & 0.19{\tiny $\pm$0.05} & 7.19{\tiny $\pm$0.96} \\

 & 0 & Verbalized & 1.76{\tiny $\pm$0.41} & 0.12{\tiny $\pm$0.03} & 0.12{\tiny $\pm$0.03} & 6.97{\tiny $\pm$0.30} & 2.56{\tiny $\pm$0.62} & 0.12{\tiny $\pm$0.04} & 0.12{\tiny $\pm$0.04} & 7.45{\tiny $\pm$0.52} \\
 & 1 & SSOT & 1.64{\tiny $\pm$0.37} & 0.09{\tiny $\pm$0.04} & 0.15{\tiny $\pm$0.04} & 6.98{\tiny $\pm$0.28} & 2.96{\tiny $\pm$0.64} & 0.19{\tiny $\pm$0.08} & 0.15{\tiny $\pm$0.04} & 7.29{\tiny $\pm$0.58} \\
\cmidrule(lr){2-11}
 & \cellcolor{blue!10} 2 & \cellcolor{blue!10} Keyword (ours) & \cellcolor{blue!10} 5.75{\tiny $\pm$1.43} & \cellcolor{blue!10} 0.27{\tiny $\pm$0.06} & \cellcolor{blue!10} 0.22{\tiny $\pm$0.04} & \cellcolor{blue!10} 7.38{\tiny $\pm$0.37} & \cellcolor{blue!10} 6.36{\tiny $\pm$0.97} & \cellcolor{blue!10} 0.29{\tiny $\pm$0.05} & \cellcolor{blue!10} 0.21{\tiny $\pm$0.04} & \cellcolor{blue!10} \textbf{8.11{\tiny $\pm$0.17}} \\
 & \cellcolor{blue!10} 2 & \cellcolor{blue!10} Div Outline (ours) & \cellcolor{blue!10} \textbf{6.08{\tiny $\pm$1.39}} & \cellcolor{blue!10} \textbf{0.31{\tiny $\pm$0.07}} & \cellcolor{blue!10} \textbf{0.25{\tiny $\pm$0.04}} & \cellcolor{blue!10} \textbf{7.42{\tiny $\pm$0.29}} & \cellcolor{blue!10} \textbf{7.21{\tiny $\pm$1.57}} & \cellcolor{blue!10} \textbf{0.29{\tiny $\pm$0.06}} & \cellcolor{blue!10} \textbf{0.23{\tiny $\pm$0.05}} & \cellcolor{blue!10} 8.04{\tiny $\pm$0.23} \\
\bottomrule
\end{tabularx}
\end{table*}

\begin{table*}[htbp]
\centering
\caption{Diversity and quality on WritingPrompts. Methods are grouped by injection level. DeepSeek-V3.2 and Qwen3-4B include a high-temperature baseline (Direct at $\tau{=}1.5$). 
\textbf{Bold} = best per column within each model. $\uparrow$ = higher is better.}
\label{tab:appendix_wp}
\small
\begin{tabularx}{\textwidth}{lll *{4}{>{\centering\arraybackslash}X}}
\toprule
 &  &  & \multicolumn{4}{c}{\textbf{WritingPrompts}} \\
\cmidrule(lr){4-7}
& \textbf{Lv.} & \textbf{Method} & LLM Div $\uparrow$ & Emb Div $\uparrow$ & Struct Div $\uparrow$ & Quality $\uparrow$\\
\midrule
\multirow{5}{*}{\rotatebox{90}{GPT-5-mini}} & 0 & Direct & 4.96{\tiny $\pm$2.38} & 0.28{\tiny $\pm$0.10} & 0.22{\tiny $\pm$0.08} & 8.59{\tiny $\pm$0.21} \\
 & 0 & Verbalized & 4.69{\tiny $\pm$1.66} & 0.25{\tiny $\pm$0.07} & 0.20{\tiny $\pm$0.06} & 8.71{\tiny $\pm$0.20} \\
 & 1 & SSOT & 3.55{\tiny $\pm$1.36} & 0.21{\tiny $\pm$0.08} & 0.17{\tiny $\pm$0.06} & 8.17{\tiny $\pm$1.50} \\
\cmidrule(lr){2-7}
 & \cellcolor{blue!10} 2 & \cellcolor{blue!10} Keyword (ours) & \cellcolor{blue!10} 6.16{\tiny $\pm$1.31} & \cellcolor{blue!10} 0.29{\tiny $\pm$0.05} & \cellcolor{blue!10} 0.23{\tiny $\pm$0.05} & \cellcolor{blue!10} \textbf{8.86{\tiny $\pm$0.14}} \\
 & \cellcolor{blue!10} 2 & \cellcolor{blue!10} Div Outline (ours) & \cellcolor{blue!10} \textbf{8.71{\tiny $\pm$1.07}} & \cellcolor{blue!10} \textbf{0.37{\tiny $\pm$0.07}} & \cellcolor{blue!10} \textbf{0.29{\tiny $\pm$0.06}} & \cellcolor{blue!10} 8.73{\tiny $\pm$0.18} \\
\midrule
\multirow{6}{*}{\rotatebox{90}{DeepSeek-V3.2}} & 0 & Direct & 5.31{\tiny $\pm$1.98} & 0.34{\tiny $\pm$0.11} & 0.21{\tiny $\pm$0.06} & 8.21{\tiny $\pm$0.37} \\
 & 0 & Direct ($\tau{=}1.5$) & 4.74{\tiny $\pm$2.10} & 0.29{\tiny $\pm$0.13} & 0.20{\tiny $\pm$0.06} & 7.91{\tiny $\pm$0.43} \\
 & 0 & Verbalized & 4.73{\tiny $\pm$1.57} & 0.29{\tiny $\pm$0.10} & 0.19{\tiny $\pm$0.05} & 8.19{\tiny $\pm$0.37} \\

 & 1 & SSOT & 3.74{\tiny $\pm$1.07} & 0.24{\tiny $\pm$0.09} & 0.17{\tiny $\pm$0.04} & 8.01{\tiny $\pm$1.00} \\
\cmidrule(lr){2-7}
 & \cellcolor{blue!10} 2 & \cellcolor{blue!10} Keyword (ours) & \cellcolor{blue!10} 6.93{\tiny $\pm$1.27} & \cellcolor{blue!10} 0.34{\tiny $\pm$0.06} & \cellcolor{blue!10} 0.23{\tiny $\pm$0.05} & \cellcolor{blue!10} \textbf{8.40{\tiny $\pm$0.27}} \\
 & \cellcolor{blue!10} 2 & \cellcolor{blue!10} Div Outline (ours) & \cellcolor{blue!10} \textbf{8.07{\tiny $\pm$1.94}} & \cellcolor{blue!10} \textbf{0.38{\tiny $\pm$0.10}} & \cellcolor{blue!10} \textbf{0.26{\tiny $\pm$0.06}} & \cellcolor{blue!10} 8.16{\tiny $\pm$1.46} \\
\midrule
\multirow{5}{*}{\rotatebox{90}{Mistral-L3}} & 0 & Direct & 2.46{\tiny $\pm$1.02} & 0.21{\tiny $\pm$0.12} & 0.12{\tiny $\pm$0.05} & 7.46{\tiny $\pm$0.47} \\
 & 0 & Verbalized & 3.26{\tiny $\pm$1.36} & 0.23{\tiny $\pm$0.09} & 0.13{\tiny $\pm$0.04} & 7.74{\tiny $\pm$0.73} \\
 & 1 & SSOT & 3.76{\tiny $\pm$1.50} & 0.20{\tiny $\pm$0.08} & 0.13{\tiny $\pm$0.04} & 7.56{\tiny $\pm$1.83} \\
\cmidrule(lr){2-7}
 & \cellcolor{blue!10} 2 & \cellcolor{blue!10} Keyword (ours) & \cellcolor{blue!10} 6.15{\tiny $\pm$1.45} & \cellcolor{blue!10} 0.29{\tiny $\pm$0.06} & \cellcolor{blue!10} 0.18{\tiny $\pm$0.04} & \cellcolor{blue!10} \textbf{8.35{\tiny $\pm$0.26}} \\
 & \cellcolor{blue!10} 2 & \cellcolor{blue!10} Div Outline (ours) & \cellcolor{blue!10} \textbf{7.86{\tiny $\pm$2.05}} & \cellcolor{blue!10} \textbf{0.34{\tiny $\pm$0.07}} & \cellcolor{blue!10} \textbf{0.20{\tiny $\pm$0.05}} & \cellcolor{blue!10} 8.13{\tiny $\pm$1.19} \\
\midrule
\multirow{7}{*}{\rotatebox{90}{Qwen3-4B}} & 0 & Direct & 1.89{\tiny $\pm$0.72} & 0.15{\tiny $\pm$0.09} & 0.11{\tiny $\pm$0.05} & 7.10{\tiny $\pm$0.86} \\
 & 0 & Direct ($\tau{=}1.5$) & 2.22{\tiny $\pm$0.81} & 0.16{\tiny $\pm$0.09} & 0.12{\tiny $\pm$0.04} & 6.79{\tiny $\pm$1.02} \\
 & 0 & Verbalized & 2.06{\tiny $\pm$0.98} & 0.14{\tiny $\pm$0.07} & 0.11{\tiny $\pm$0.04} & 6.88{\tiny $\pm$0.73} \\
 & 0 & G2  &  2.00{\tiny $\pm$0.69} & 0.09{\tiny $\pm$0.03} & 0.22{\tiny $\pm$0.05} & 6.81{\tiny $\pm$1.01} \\

 & 1 & SSOT & 2.12{\tiny $\pm$0.83} & 0.14{\tiny $\pm$0.07} & 0.15{\tiny $\pm$0.05} & 6.56{\tiny $\pm$0.89} \\
\cmidrule(lr){2-7}
 & \cellcolor{blue!10} 2 & \cellcolor{blue!10} Keyword (ours) & \cellcolor{blue!10} 5.74{\tiny $\pm$1.57} & \cellcolor{blue!10} 0.30{\tiny $\pm$0.05} & \cellcolor{blue!10} 0.20{\tiny $\pm$0.05} & \cellcolor{blue!10} \textbf{8.19{\tiny $\pm$0.21}} \\
 & \cellcolor{blue!10} 2 & \cellcolor{blue!10} Div Outline (ours) & \cellcolor{blue!10} \textbf{6.51{\tiny $\pm$1.80}} & \cellcolor{blue!10} \textbf{0.34{\tiny $\pm$0.06}} & \cellcolor{blue!10} \textbf{0.22{\tiny $\pm$0.06}} & \cellcolor{blue!10} 8.04{\tiny $\pm$0.29} \\
\bottomrule
\end{tabularx}
\end{table*}

\begin{table*}[t]
\centering
\caption{Lexical diversity (Distinct-3 and Self-BLEU) on Infinite Chats and NoveltyBench. Methods are grouped by injection level. DeepSeek-V3.2 and Qwen3-4B include a high-temperature baseline (Direct at $\tau{=}1.5$). \textbf{Bold} = best per column within each model. $\uparrow$ = higher is better, $\downarrow$ = lower is better.}
\label{tab:lexical_div_main}
\small
\begin{tabularx}{\textwidth}{lll *{2}{>{\centering\arraybackslash}X} *{2}{>{\centering\arraybackslash}X}}
\toprule
 &  &  & \multicolumn{2}{c}{\textbf{Infinite Chats}} & \multicolumn{2}{c}{\textbf{NoveltyBench}} \\
\cmidrule(lr){4-5} \cmidrule(lr){6-7}
& \textbf{Lv.} & \textbf{Method} & Distinct-3 $\uparrow$ & Self-BLEU $\downarrow$ & Distinct-3 $\uparrow$ & Self-BLEU $\downarrow$\\
\midrule
\multirow{5}{*}{\rotatebox{90}{GPT-5-mini}} & 0 & Direct & 0.87{\tiny $\pm$0.09} & 0.26{\tiny $\pm$0.14} & 0.89{\tiny $\pm$0.07} & 0.23{\tiny $\pm$0.11} \\
 & 0 & Verbalized & 0.88{\tiny $\pm$0.07} & 0.26{\tiny $\pm$0.11} & 0.87{\tiny $\pm$0.06} & 0.28{\tiny $\pm$0.10} \\
 & 1 & SSOT & 0.78{\tiny $\pm$0.11} & 0.38{\tiny $\pm$0.14} & 0.76{\tiny $\pm$0.11} & 0.42{\tiny $\pm$0.15} \\
\cmidrule(lr){2-7}
 & \cellcolor{blue!10} 2 & \cellcolor{blue!10} Keyword (ours) & \cellcolor{blue!10} 0.91{\tiny $\pm$0.07} & \cellcolor{blue!10} 0.21{\tiny $\pm$0.10} & \cellcolor{blue!10} 0.92{\tiny $\pm$0.04} & \cellcolor{blue!10} 0.17{\tiny $\pm$0.07} \\
 & \cellcolor{blue!10} 2 & \cellcolor{blue!10} Div Outline (ours) & \cellcolor{blue!10} \textbf{0.94{\tiny $\pm$0.04}} & \cellcolor{blue!10} \textbf{0.12{\tiny $\pm$0.07}} & \cellcolor{blue!10} \textbf{0.95{\tiny $\pm$0.04}} & \cellcolor{blue!10} \textbf{0.11{\tiny $\pm$0.06}} \\
\midrule
\multirow{6}{*}{\rotatebox{90}{DeepSeek-V3.2}} & 0 & Direct & 0.91{\tiny $\pm$0.05} & 0.19{\tiny $\pm$0.09} & 0.90{\tiny $\pm$0.06} & 0.21{\tiny $\pm$0.09} \\
 & 0 & Direct ($\tau{=}1.5$) & \textbf{0.98{\tiny $\pm$0.04}} & \textbf{0.06{\tiny $\pm$0.05}} & \textbf{0.97{\tiny $\pm$0.10}} & \textbf{0.07{\tiny $\pm$0.05}} \\
 & 0 & Verbalized & 0.85{\tiny $\pm$0.08} & 0.31{\tiny $\pm$0.11} & 0.82{\tiny $\pm$0.12} & 0.34{\tiny $\pm$0.14} \\

 & 1 & SSOT & 0.77{\tiny $\pm$0.12} & 0.40{\tiny $\pm$0.15} & 0.78{\tiny $\pm$0.14} & 0.37{\tiny $\pm$0.13} \\
\cmidrule(lr){2-7}
 & \cellcolor{blue!10} 2 & \cellcolor{blue!10} Keyword (ours) & \cellcolor{blue!10} 0.92{\tiny $\pm$0.06} & \cellcolor{blue!10} 0.17{\tiny $\pm$0.09} & \cellcolor{blue!10} 0.94{\tiny $\pm$0.04} & \cellcolor{blue!10} 0.14{\tiny $\pm$0.06} \\
 & \cellcolor{blue!10} 2 & \cellcolor{blue!10} Div Outline (ours) & \cellcolor{blue!10} 0.93{\tiny $\pm$0.05} & \cellcolor{blue!10} 0.16{\tiny $\pm$0.09} & \cellcolor{blue!10} 0.92{\tiny $\pm$0.09} & \cellcolor{blue!10} 0.15{\tiny $\pm$0.07} \\
\midrule
\multirow{5}{*}{\rotatebox{90}{Mistral-L3}} & 0 & Direct & 0.62{\tiny $\pm$0.23} & 0.57{\tiny $\pm$0.27} & 0.70{\tiny $\pm$0.17} & 0.46{\tiny $\pm$0.20} \\
 & 0 & Verbalized & 0.71{\tiny $\pm$0.12} & 0.49{\tiny $\pm$0.13} & 0.70{\tiny $\pm$0.13} & 0.50{\tiny $\pm$0.14} \\
 & 1 & SSOT & 0.71{\tiny $\pm$0.11} & 0.48{\tiny $\pm$0.13} & 0.66{\tiny $\pm$0.15} & 0.52{\tiny $\pm$0.16} \\
\cmidrule(lr){2-7}
 & \cellcolor{blue!10} 2 & \cellcolor{blue!10} Keyword (ours) & \cellcolor{blue!10} 0.89{\tiny $\pm$0.05} & \cellcolor{blue!10} 0.22{\tiny $\pm$0.08} & \cellcolor{blue!10} \textbf{0.91{\tiny $\pm$0.03}} & \cellcolor{blue!10} 0.18{\tiny $\pm$0.05} \\
 & \cellcolor{blue!10} 2 & \cellcolor{blue!10} Div Outline (ours) & \cellcolor{blue!10} \textbf{0.91{\tiny $\pm$0.05}} & \cellcolor{blue!10} \textbf{0.19{\tiny $\pm$0.08}} & \cellcolor{blue!10} 0.91{\tiny $\pm$0.09} & \cellcolor{blue!10} \textbf{0.17{\tiny $\pm$0.06}} \\
\midrule
\multirow{6}{*}{\rotatebox{90}{Qwen3-4B}} & 0 & Direct & 0.79{\tiny $\pm$0.12} & 0.37{\tiny $\pm$0.15} & 0.78{\tiny $\pm$0.13} & 0.37{\tiny $\pm$0.14} \\
 & 0 & Direct ($\tau{=}1.5$) & 0.86{\tiny $\pm$0.11} & 0.28{\tiny $\pm$0.15} & 0.81{\tiny $\pm$0.13} & 0.34{\tiny $\pm$0.15} \\
 & 0 & Verbalized & 0.78{\tiny $\pm$0.10} & 0.40{\tiny $\pm$0.13} & 0.77{\tiny $\pm$0.18} & 0.36{\tiny $\pm$0.13} \\
 & 0 & G2 & 0.71{\tiny $\pm$0.11} & 0.47{\tiny $\pm$0.13} & 0.80{\tiny $\pm$0.06} & 0.37{\tiny $\pm$0.08} \\

 & 1 & SSOT & 0.67{\tiny $\pm$0.12} & 0.51{\tiny $\pm$0.13} & 0.68{\tiny $\pm$0.12} & 0.51{\tiny $\pm$0.14} \\
\cmidrule(lr){2-7}
 & \cellcolor{blue!10} 2 & \cellcolor{blue!10} Keyword (ours) & \cellcolor{blue!10} 0.88{\tiny $\pm$0.07} & \cellcolor{blue!10} \textbf{0.23{\tiny $\pm$0.08}} & \cellcolor{blue!10} 0.87{\tiny $\pm$0.10} & \cellcolor{blue!10} 0.24{\tiny $\pm$0.07} \\
 & \cellcolor{blue!10} 2 & \cellcolor{blue!10} Div Outline (ours) & \cellcolor{blue!10} \textbf{0.88{\tiny $\pm$0.06}} & \cellcolor{blue!10} 0.23{\tiny $\pm$0.09} & \cellcolor{blue!10} \textbf{0.88{\tiny $\pm$0.06}} & \cellcolor{blue!10} \textbf{0.24{\tiny $\pm$0.09}} \\
\bottomrule
\end{tabularx}
\end{table*}

\begin{table*}[t]
\centering
\caption{Lexical diversity (Distinct-3 and Self-BLEU) on WritingPrompts, IBM ArgKP, and Poetry Foundation. Methods are grouped by injection level. DeepSeek-V3.2 and Qwen3-4B include a high-temperature baseline (Direct at $\tau{=}1.5$). \textbf{Bold} = best per column within each model. $\uparrow$ = higher is better, $\downarrow$ = lower is better.}
\label{tab:lexical_div_appendix}
\small
\begin{tabularx}{\textwidth}{lll *{2}{>{\centering\arraybackslash}X} *{2}{>{\centering\arraybackslash}X} *{2}{>{\centering\arraybackslash}X}}
\toprule
 &  &  & \multicolumn{2}{c}{\textbf{WritingPrompts}} & \multicolumn{2}{c}{\textbf{IBM ArgKP}} & \multicolumn{2}{c}{\textbf{Poetry Foundation}} \\
\cmidrule(lr){4-5} \cmidrule(lr){6-7} \cmidrule(lr){8-9}
& \textbf{Lv.} & \textbf{Method} & Distinct-3 $\uparrow$ & Self-BLEU $\downarrow$ & Distinct-3 $\uparrow$ & Self-BLEU $\downarrow$ & Distinct-3 $\uparrow$ & Self-BLEU $\downarrow$\\
\midrule
\multirow{5}{*}{\rotatebox{90}{GPT-5-mini}} & 0 & Direct & 0.92{\tiny $\pm$0.07} & 0.18{\tiny $\pm$0.11} & 0.94{\tiny $\pm$0.02} & 0.16{\tiny $\pm$0.04} & 0.88{\tiny $\pm$0.04} & 0.25{\tiny $\pm$0.05} \\
 & 0 & Verbalized & 0.92{\tiny $\pm$0.05} & 0.18{\tiny $\pm$0.09} & 0.91{\tiny $\pm$0.02} & 0.23{\tiny $\pm$0.04} & \textbf{0.90{\tiny $\pm$0.03}} & 0.22{\tiny $\pm$0.05} \\
 & 1 & SSOT & 0.84{\tiny $\pm$0.16} & 0.27{\tiny $\pm$0.11} & 0.81{\tiny $\pm$0.03} & 0.38{\tiny $\pm$0.04} & 0.88{\tiny $\pm$0.03} & 0.25{\tiny $\pm$0.05} \\
\cmidrule(lr){2-9}
 & \cellcolor{blue!10} 2 & \cellcolor{blue!10} Keyword (ours) & \cellcolor{blue!10} 0.94{\tiny $\pm$0.04} & \cellcolor{blue!10} 0.14{\tiny $\pm$0.06} & \cellcolor{blue!10} 0.91{\tiny $\pm$0.01} & \cellcolor{blue!10} 0.22{\tiny $\pm$0.02} & \cellcolor{blue!10} 0.85{\tiny $\pm$0.09} & \cellcolor{blue!10} 0.16{\tiny $\pm$0.05} \\
 & \cellcolor{blue!10} 2 & \cellcolor{blue!10} Div Outline (ours) & \cellcolor{blue!10} \textbf{0.95{\tiny $\pm$0.03}} & \cellcolor{blue!10} \textbf{0.10{\tiny $\pm$0.05}} & \cellcolor{blue!10} \textbf{0.96{\tiny $\pm$0.01}} & \cellcolor{blue!10} \textbf{0.11{\tiny $\pm$0.03}} & \cellcolor{blue!10} 0.88{\tiny $\pm$0.09} & \cellcolor{blue!10} \textbf{0.14{\tiny $\pm$0.05}} \\
\midrule
\multirow{6}{*}{\rotatebox{90}{DeepSeek-V3.2}} & 0 & Direct & 0.93{\tiny $\pm$0.04} & 0.17{\tiny $\pm$0.07} & 0.95{\tiny $\pm$0.02} & 0.13{\tiny $\pm$0.04} & 0.91{\tiny $\pm$0.05} & 0.21{\tiny $\pm$0.08} \\
& 0 & Direct ($\tau{=}1.5$) & \textbf{0.98{\tiny $\pm$0.02}} & \textbf{0.07{\tiny $\pm$0.05}} & \textbf{0.98{\tiny $\pm$0.01}} & \textbf{0.06{\tiny $\pm$0.03}} & \textbf{0.97{\tiny $\pm$0.02}} & \textbf{0.09{\tiny $\pm$0.05}} \\
 & 0 & Verbalized & 0.91{\tiny $\pm$0.04} & 0.20{\tiny $\pm$0.07} & 0.89{\tiny $\pm$0.02} & 0.28{\tiny $\pm$0.04} & 0.91{\tiny $\pm$0.06} & 0.21{\tiny $\pm$0.07} \\
 & 1 & SSOT & 0.87{\tiny $\pm$0.05} & 0.28{\tiny $\pm$0.08} & 0.78{\tiny $\pm$0.04} & 0.43{\tiny $\pm$0.05} & 0.86{\tiny $\pm$0.04} & 0.28{\tiny $\pm$0.04} \\
\cmidrule(lr){2-9}
 & \cellcolor{blue!10} 2 & \cellcolor{blue!10} Keyword (ours) & \cellcolor{blue!10} 0.93{\tiny $\pm$0.03} & \cellcolor{blue!10} 0.17{\tiny $\pm$0.05} & \cellcolor{blue!10} 0.92{\tiny $\pm$0.02} & \cellcolor{blue!10} 0.20{\tiny $\pm$0.04} & \cellcolor{blue!10} 0.87{\tiny $\pm$0.08} & \cellcolor{blue!10} 0.17{\tiny $\pm$0.05} \\
 & \cellcolor{blue!10} 2 & \cellcolor{blue!10} Div Outline (ours) & \cellcolor{blue!10} 0.91{\tiny $\pm$0.16} & \cellcolor{blue!10} 0.13{\tiny $\pm$0.05} & \cellcolor{blue!10} 0.94{\tiny $\pm$0.01} & \cellcolor{blue!10} 0.16{\tiny $\pm$0.03} & \cellcolor{blue!10} 0.90{\tiny $\pm$0.04} & \cellcolor{blue!10} 0.13{\tiny $\pm$0.05} \\
\midrule
\multirow{5}{*}{\rotatebox{90}{Mistral-L3}} & 0 & Direct & 0.51{\tiny $\pm$0.19} & 0.70{\tiny $\pm$0.22} & 0.69{\tiny $\pm$0.20} & 0.50{\tiny $\pm$0.25} & 0.35{\tiny $\pm$0.11} & 0.89{\tiny $\pm$0.12} \\
 & 0 & Verbalized & 0.79{\tiny $\pm$0.08} & 0.40{\tiny $\pm$0.10} & 0.74{\tiny $\pm$0.06} & 0.48{\tiny $\pm$0.07} & 0.80{\tiny $\pm$0.04} & 0.38{\tiny $\pm$0.06} \\
 & 1 & SSOT & 0.77{\tiny $\pm$0.18} & 0.37{\tiny $\pm$0.09} & 0.69{\tiny $\pm$0.04} & 0.52{\tiny $\pm$0.05} & 0.81{\tiny $\pm$0.04} & 0.35{\tiny $\pm$0.05} \\
\cmidrule(lr){2-9}
 & \cellcolor{blue!10} 2 & \cellcolor{blue!10} Keyword (ours) & \cellcolor{blue!10} 0.88{\tiny $\pm$0.03} & \cellcolor{blue!10} 0.24{\tiny $\pm$0.05} & \cellcolor{blue!10} 0.89{\tiny $\pm$0.02} & \cellcolor{blue!10} 0.24{\tiny $\pm$0.04} & \cellcolor{blue!10} 0.83{\tiny $\pm$0.03} & \cellcolor{blue!10} 0.27{\tiny $\pm$0.04} \\
 & \cellcolor{blue!10} 2 & \cellcolor{blue!10} Div Outline (ours) & \cellcolor{blue!10} \textbf{0.89{\tiny $\pm$0.13}} & \cellcolor{blue!10} \textbf{0.18{\tiny $\pm$0.05}} & \cellcolor{blue!10} \textbf{0.91{\tiny $\pm$0.02}} & \cellcolor{blue!10} \textbf{0.18{\tiny $\pm$0.03}} & \cellcolor{blue!10} \textbf{0.87{\tiny $\pm$0.03}} & \cellcolor{blue!10} \textbf{0.22{\tiny $\pm$0.04}} \\
\midrule
\multirow{7}{*}{\rotatebox{90}{Qwen3-4B}} & 0 & Direct & 0.79{\tiny $\pm$0.08} & 0.39{\tiny $\pm$0.10} & 0.85{\tiny $\pm$0.03} & 0.32{\tiny $\pm$0.06} & 0.80{\tiny $\pm$0.06} & 0.37{\tiny $\pm$0.07} \\
& 0 & Direct ($\tau{=}1.5$) & 0.83{\tiny $\pm$0.10} & 0.33{\tiny $\pm$0.13} & 0.87{\tiny $\pm$0.10} & 0.26{\tiny $\pm$0.16} & \textbf{0.89{\tiny $\pm$0.05}} & 0.23{\tiny $\pm$0.07} \\
 & 0 & Verbalized & 0.78{\tiny $\pm$0.08} & 0.40{\tiny $\pm$0.09} & 0.82{\tiny $\pm$0.03} & 0.36{\tiny $\pm$0.03} & 0.84{\tiny $\pm$0.05} & 0.32{\tiny $\pm$0.07} \\
  & 0 & G2 & 0.71{\tiny $\pm$0.09} & 0.48{\tiny $\pm$0.10} & 0.70{\tiny $\pm$0.04} & 0.51{\tiny $\pm$0.04} & 0.80{\tiny $\pm$0.06} & 0.37{\tiny $\pm$0.07}  \\

 & 1 & SSOT & 0.71{\tiny $\pm$0.08} & 0.47{\tiny $\pm$0.09} & 0.71{\tiny $\pm$0.04} & 0.50{\tiny $\pm$0.04} & 0.82{\tiny $\pm$0.09} & 0.29{\tiny $\pm$0.05} \\
\cmidrule(lr){2-9}
 & \cellcolor{blue!10} 2 & \cellcolor{blue!10} Keyword (ours) & \cellcolor{blue!10} 0.89{\tiny $\pm$0.03} & \cellcolor{blue!10} 0.22{\tiny $\pm$0.04} & \cellcolor{blue!10} \textbf{0.89{\tiny $\pm$0.03}} & \cellcolor{blue!10} \textbf{0.23{\tiny $\pm$0.05}} & \cellcolor{blue!10} 0.89{\tiny $\pm$0.04} & \cellcolor{blue!10} 0.19{\tiny $\pm$0.05} \\
 & \cellcolor{blue!10} 2 & \cellcolor{blue!10} Div Outline (ours) & \cellcolor{blue!10} \textbf{0.90{\tiny $\pm$0.03}} & \cellcolor{blue!10} \textbf{0.20{\tiny $\pm$0.05}} & \cellcolor{blue!10} 0.89{\tiny $\pm$0.02} & \cellcolor{blue!10} 0.23{\tiny $\pm$0.04} & \cellcolor{blue!10} 0.89{\tiny $\pm$0.04} & \cellcolor{blue!10} \textbf{0.19{\tiny $\pm$0.05}} \\
\bottomrule
\end{tabularx}
\end{table*}

\begin{table}[htbp]
\centering
\caption{Transmission score $T$, realized diversity $I(Y;Z\mid X)$, and output entropy $H(Y\mid X)$ on Qwen3-4B for WritingPrompts, Poetry Foundation, and IBM ArgKP. $T$ and $I(Y;Z\mid X)$ are undefined for Level~0 methods. Level~2 methods consistently achieve higher $T$ and $I(Y;Z\mid X)$ than the Level~1 baseline.}
\label{tab:T_values_appendix}
\small
\setlength{\tabcolsep}{2.5pt}
\begin{tabular}{llccc}
\toprule
Data & Method & $T$ $\uparrow$ & $I(Y;Z{\mid}X)$ $\uparrow$ & $H(Y{\mid}X)$ $\uparrow$ \\
\midrule
\multirow{3}{*}{\rotatebox{90}{\shortstack{Writing\\Prompts}}}
& \cellcolor{blue!10} Div Outline (ours) & \cellcolor{blue!10} \textbf{0.463} & \cellcolor{blue!10} 1.46 & \cellcolor{blue!10} 3.08 \\
& \cellcolor{blue!10} Keyword (ours)     & \cellcolor{blue!10} 0.388          & \cellcolor{blue!10} 1.12 & \cellcolor{blue!10} 2.85 \\
& SSOT                                   & 0.003                              & 0.00                     & 0.64 \\
\midrule
\multirow{3}{*}{\rotatebox{90}{\shortstack{Poetry\\Found.}}}
& \cellcolor{blue!10} Div Outline (ours) & \cellcolor{blue!10} \textbf{0.243} & \cellcolor{blue!10} 0.94 & \cellcolor{blue!10} 3.86 \\
& \cellcolor{blue!10} Keyword (ours)     & \cellcolor{blue!10} 0.165          & \cellcolor{blue!10} 0.65 & \cellcolor{blue!10} 3.94 \\
& SSOT                                   & 0.000                              & 0.00                  & 1.22 \\
\midrule
\multirow{3}{*}{\rotatebox{90}{\shortstack{IBM\\ArgKP}}}
& \cellcolor{blue!10} Div Outline (ours) & \cellcolor{blue!10} \textbf{0.565} & \cellcolor{blue!10} 1.61 & \cellcolor{blue!10} 2.82 \\
& \cellcolor{blue!10} Keyword (ours)     & \cellcolor{blue!10} 0.457          & \cellcolor{blue!10} 1.25 & \cellcolor{blue!10} 2.70 \\
& SSOT                                   & 0.002                              & 0.00                     & 0.82 \\
\bottomrule
\end{tabular}
\end{table}

\subsection{Experiment Prompts and Example Outputs}
\label{sec:appendix_prompt_and_result}
\subsubsection{Diverse Outline: Prompts and Generation Details}
\label{sec:appendix_no_audit_icl}

As described in Section \ref{sec:method}, Diverse Outline generates all $m$ specifications in a single LLM call, then conditions on each to produce one output. We provide the prompt details for both steps below.

\paragraph{Outline generation.}
Each specification $z_i$ takes the form of a compact keyword outline: a list of 4-6 keywords capturing tone, format, perspective, and key focus. The system prompt instructs the model to produce such outlines simultaneously, requiring that no two outlines share more than one keyword. Because the model sees its own previous outlines while generating each subsequent one, it naturally enforces diversity across the full set. 
The full prompt is shown in Figure~\ref{fig:noaudit_gen_prompt}.

\begin{figure}[htbp]
\small\centering
\begin{minipage}{0.95\linewidth}
\noindent\rule{\linewidth}{0.4pt}\\[2pt]
\textbf{System Prompt --- Diverse Outline Generation}\\[4pt]
You are a diversity planning engine.
Goal: produce \{N\} DISTINCT keyword outlines for the SAME request.\\[4pt]
Rules:
\begin{enumerate}[leftmargin=*,nosep]
  \item Each outline is a compact list of 4--6 keywords/phrases capturing: tone, format, perspective, and key focus.
        Good examples: ``instructional tone'', ``bullet-point format'', ``second-person'', ``focus on failure cases''.
  \item Maximize diversity: no two outlines should share more than 1 keyword.
  \item Keywords must be task-specific --- encode what makes each approach different for this particular request.
\end{enumerate}
\vspace{4pt}
Output JSON only:
\begin{verbatim}
{
  "outlines": [
    { "id": 1, 
    "keywords": ["keyword1", ..., "keyword4"] },
    ...
  ]
}
\end{verbatim}
\textbf{User Prompt}\\[4pt]
Task: \{TASK\}\\[2pt]
\noindent\rule{\linewidth}{0.4pt}
\end{minipage}
\caption{Prompt used to generate $m$ diverse keyword outlines in one batch call in Diverse Outline.}
\label{fig:noaudit_gen_prompt}
\end{figure}

\paragraph{Output generation.}
Each outline is then used to condition the generation of one output.
The model is guided by the keyword list but not rigidly required to reproduce every keyword verbatim; the keywords shape the overall tone, format, and focus of the response. The generation prompt is shown in Figure~\ref{fig:noaudit_story_prompt}.

\begin{figure}[htbp]
\small\centering
\begin{minipage}{0.95\linewidth}
\noindent\rule{\linewidth}{0.4pt}\\[2pt]
\textbf{System Prompt --- Output Generation (Diverse Outline)}\\[4pt]
You are a helpful AI assistant.
Your task is to respond to the given task, guided by the provided keywords.\\[4pt]
\emph{Input:}
\begin{itemize}[leftmargin=*,nosep]
  \item \textbf{Task:} the original request (follow ALL constraints in it, e.g.\ length, format, topic).
  \item \textbf{Outline:} a JSON object with axis:value pairs describing structural choices (e.g.\ narrative\_pov, tone, structure, theme).
\end{itemize}
\vspace{4pt}
\emph{Instructions:}
\begin{enumerate}[leftmargin=*,nosep]
  \item Strictly follow any constraints in the Task.
  \item Let each keyword shape the tone, format, and focus of your response.
  \item Keep your response to approximately 200 words.
  \item Do not list or explain the keywords. Output ONLY the final response text.
\end{enumerate}
\vspace{2pt}
\noindent\rule{\linewidth}{0.4pt}
\end{minipage}
\caption{Prompt used to generate outputs conditioned on a Diverse Outline specification.}
\label{fig:noaudit_story_prompt}
\end{figure}

\subsubsection{Keyword: Axis Generation and Combination Strategy}
\label{sec:appendix_keyword_combo}
As described in Section \ref{sec:method}, Keyword first generates a set of dimensions and candidate values, then selects $m$ combinations to construct specifications. We provide the prompt and selection details below.

\paragraph{Axis generation.}
For each prompt, we generate $A=4$ dimensions (referred to as axes in the prompt) and $V=8$ values each in a single call. 
Axes are required to be orthogonal: selecting a value on one axis should not constrain choices on other axes. The dimensions are generated once per prompt and shared across all outputs. The full prompt is shown in Figure \ref{fig:axis_gen_prompt}.

\begin{figure}[htbp]
\small\centering
\begin{minipage}{0.95\linewidth}
\noindent\rule{\linewidth}{0.4pt}\\[2pt]
\textbf{System Prompt — Axis Generation}\\[4pt]
You are a creative diversity planner.
Generate \{N\_KEYS\} independent structural dimensions (axes) that capture the most impactful
creative choices for responses to the given task.
For each axis, generate \{N\_VALUES\} distinct values representing meaningfully different choices.
Axes must be orthogonal --- choosing a value for one axis should not constrain choices for other axes.\\[4pt]
Output JSON only:
\begin{verbatim}
{
  "axes": [
    {
      "key": "narrative_pov",
      "label": "Narrative POV",
      "values": ["first-person unreliable",
                 "third-person omniscient", ...]
    }, ...
  ]
}
\end{verbatim}
\textbf{User Prompt}\\[4pt]
Task type: \{TASK\_TYPE\}\quad Task: \{TASK\}\\
Generate \{N\_KEYS\} axes with \{N\_VALUES\} values each.\\[2pt]
\noindent\rule{\linewidth}{0.4pt}
\end{minipage}
\caption{Prompt used to generate diversity axes for Keyword. We use $A=4$ axes and $V=8$ values per axis.}
\label{fig:axis_gen_prompt}
\end{figure}

\paragraph{Combination selection.}
Given $A$ axes each with $V$ values, there are $V^A = 8^4 = 4{,}096$ possible axis-value combinations.
We select $m=20$ combinations using a \emph{greedy Hamming-distance maximization} strategy.

Formally, represent each combination as a tuple $\mathbf{c} = (v_1, v_2, v_3, v_4) \in \{1,\ldots,V\}^A$.
The Hamming distance between two combinations $\mathbf{c}$ and $\mathbf{c}'$ is the number of axes on which they differ:
\[
  d_H(\mathbf{c}, \mathbf{c}') = \sum_{a=1}^{A} \mathbf{1}[v_a \neq v'_a].
\]
Two combinations with $d_H = A$ differ on every axis; two with $d_H = 0$ are identical.

The algorithm proceeds greedily:
\begin{enumerate}[leftmargin=*,nosep]
  \item Draw a random combination $\mathbf{c}_1$ uniformly from the full space as the seed.
  \item For $i = 2, \ldots, m$: select the candidate $\mathbf{c}^*$ that maximizes its \emph{minimum} Hamming distance to all already-selected combinations,
  \[
    \mathbf{c}_i = \arg\max_{\mathbf{c} \notin \mathcal{S}} \min_{\mathbf{c}' \in \mathcal{S}}\, d_H(\mathbf{c}, \mathbf{c}'),
  \]
  where $\mathcal{S} = \{\mathbf{c}_1, \ldots, \mathbf{c}_{i-1}\}$ is the set of already-selected combinations.
  Ties are broken randomly.
\end{enumerate}
This max-min criterion acts as a greedy farthest-point sampler in Hamming space, ensuring that every newly added combination is as different as possible from all existing ones on as many axes as possible.
For $m=20$ and $A=4$, the resulting set typically achieves a minimum pairwise Hamming distance of $\geq 2$, meaning no two selected outlines agree on more than $A-2 = 2$ axes simultaneously.
\paragraph{Output generation.}
Each of the $m$ selected combinations is used as a structured outline to condition one output.
The generation prompt is shown in Figure~\ref{fig:keyword_gen_prompt}.
The model is explicitly instructed that every axis value must be clearly and visibly present in the generated output.

\begin{figure}[htbp]
\small\centering
\begin{minipage}{0.95\linewidth}
\noindent\rule{\linewidth}{0.4pt}\\[2pt]
\textbf{System Prompt --- Output Generation (Keyword)}\\[4pt]
You are a helpful AI assistant.
Your task is to respond to the given task, strictly manifesting the structural choices in the Outline.\\[4pt]
\emph{Input:}
\begin{itemize}[leftmargin=*,nosep]
    \item \textbf{Task:} the original request (follow ALL constraints in it, e.g.\ length, format, topic).
    \item \textbf{Outline:} a JSON object with axis:value pairs that you MUST embody (e.g.\ narrative\_pov, tone, structure, theme).
\end{itemize}
\vspace{4pt}
\emph{Instructions:}
\begin{enumerate}[leftmargin=*,nosep]
    \item Strictly follow any constraints in the Task.
    \item EACH axis value in the Outline MUST be clearly and visibly present in your response.
    A reader should be able to identify every axis value directly from your text without being told.
    \item Keep your response to approximately 200 words.
    \item Do not list or explain the axis values. Output ONLY the final response text.
\end{enumerate}
\vspace{2pt}
\noindent\rule{\linewidth}{0.4pt}
\end{minipage}
\caption{Prompt used to generate outputs conditioned on a Keyword specification.}
\label{fig:keyword_gen_prompt}
\end{figure}

\begin{table*}[t]
\renewcommand{\arraystretch}{1.35}
\setlength{\tabcolsep}{5pt}
\centering
\caption{%
  Example outputs from GPT-5-mini on a question
  \textit{``Give one popular name for a baby boy in Papua New Guinea.''}.
  Each row shows one representative output; the group label in brackets is the equivalence class
  assigned by the evaluator.
  The Direct baseline produces outputs all following the template
  \textit{``[English Christian name] is popular due to missionary influence\ldots''}
  Both Level-2 methods produce outputs differing in name choice,
  language variety (English, Tok Pisin, Hiri Motu, oral-tradition style), and regional framing.%
}
\label{fig:example_outputs}
\small
\begin{tabular}{LLL}
\toprule
\multicolumn{1}{>{\centering\arraybackslash\small}p{0.305\textwidth}}{\cellcolor{directbg}\textbf{Direct (Baseline)}}
&
\multicolumn{1}{>{\centering\arraybackslash\small}p{0.305\textwidth}}{\cellcolor{divbg}\textbf{Diverse Outline (ours)}}
&
\multicolumn{1}{>{\centering\arraybackslash\small}p{0.305\textwidth}}{\cellcolor{kwbg}\textbf{Keyword Combo (ours)}}
\\

\midrule

\cellcolor{directbg}{}
\textbf{John} is a widely used and familiar name for baby boys in Papua New Guinea. Due to the strong influence of Christian missionaries and the legacy of English, John has become one of the most common names\ldots
&
\cellcolor{divbg}{\color{gray} \textit{(informal, English)}}\newline
Try: \textbf{John} --- ``God is gracious.'' This is a friendly, simple name you'll see often on newborn lists from Lae urban hospitals. It works well in both English and Tok Pisin.
&
\cellcolor{kwbg}{\color{gray}\textit{(modern global name, island provinces)}}\newline
\textbf{Michael} --- Modern global / borrowed name. Popular baby-boy choice in Papua New Guinea's island provinces, often adopted by Christian families from Manus and New Ireland.
\\[4pt]

\cellcolor{directbg}{}
\textbf{Michael} is a popular and recognizable name for baby boys in Papua New Guinea, reflecting the ongoing cultural influence of Christianity and English-language traditions across the country\ldots
&
\cellcolor{divbg}{\color{gray} \textit{(authoritative, Papuan Malay register)}}\newline
Recommendation: \textbf{Johan}. From the Oro Province provincial census baby name sample, Johan is one of the top-five recorded male given names for the 2019--2023 cohort. Its Papuan Malay origin distinguishes it from the more common anglicised John.
&
\cellcolor{kwbg}{\color{gray}\textit{(Austronesian origin, Milne Bay)}}\newline
\textbf{Manu} --- Austronesian / island-language origin. A beloved choice from the islands, heard in canoe-launching ceremonies and school enrollment registers along the Milne Bay coast.
\\[4pt]

\cellcolor{directbg}{}
\textbf{David} is a common and well-known name for baby boys in Papua New Guinea, frequently chosen by parents influenced by Christian naming traditions and English cultural norms\ldots
&
\cellcolor{divbg}{\color{gray} \textit{(sincere, island dialect style)}}\newline
One very common name you will meet across Papua New Guinea is \textbf{Peter} --- often heard and written as \textit{Pita} in local island dialect phrasing. Unlike the more urbanised John, Peter carries a seafaring, coastal resonance in communities along the Bismarck Archipelago.
&
\cellcolor{kwbg}{\color{gray} \textit{(Trans-New Guinea highlands origin)}}\newline
\textbf{Kumul} --- Trans-New Guinea / highlands-language origin. In Port Moresby schools, this name evokes pride in the Bird of Paradise, PNG's national emblem. Parents from Simbu and Western Highlands provinces favour it for its cultural resonance.
\\[4pt]

\cellcolor{directbg}{}
\textbf{Paul} is a commonly used name for baby boys in Papua New Guinea, reflecting the strong presence of Christian traditions and English-language influence across the country\ldots
&
\cellcolor{divbg}{\color{gray} \textit{(traditional, oral-narrative style)}}\newline
Elders on the seaside speak this name like a blessing passed from one canoe to the next: \textbf{Devid} --- a name that arrived with the first mission ship and took root in coastal villages. The local spelling marks it as no longer foreign.
&
\cellcolor{kwbg}{\color{gray} \textit{(local clan / ceremonial name, Madang)}}\newline
\textbf{Kambu} --- a local clan or traditional ceremonial name heard around Madang and northern coastal communities. Unlike borrowed Christian names, Kambu carries ancestral lineage meaning and is given during a formal naming ceremony.
\\
\bottomrule
\end{tabular}
\end{table*}
\begin{table*}[t]
\renewcommand{\arraystretch}{1.35}
\setlength{\tabcolsep}{5pt}
\centering
\caption{%
  Example outputs from GPT-5-mini on question
  \textit{``What is the single best first phone for a teenager?
  You must provide only one phone as your recommendation.''}.
  The Direct baseline recommends \textbf{Google Pixel 7a} across all 20 outputs.
  Level-2 methods recommend different devices from different expert perspectives,
  varying in recommended phone, primary criterion, and response format.%
}
\label{fig:example_outputs_phone}
\small
\begin{tabular}{LLL}
\toprule
\multicolumn{1}{>{\centering\arraybackslash\small}p{0.305\textwidth}}{\cellcolor{directbg}\textbf{Direct (Baseline)}}
&
\multicolumn{1}{>{\centering\arraybackslash\small}p{0.305\textwidth}}{\cellcolor{divbg}\textbf{Diverse Outline (ours)}}
&
\multicolumn{1}{>{\centering\arraybackslash\small}p{0.305\textwidth}}{\cellcolor{kwbg}\textbf{Keyword Combo (ours)}}
\\
\midrule

\cellcolor{directbg}{}
If I had to pick a single best first phone for a teenager, I'd recommend the \textbf{Google Pixel 7a}. It strikes an excellent balance of price, performance, and useful features that matter for a teenager's day-to-day life\ldots
&
\cellcolor{divbg}{\color{gray}\textit{(parent-to-parent, parental controls)}}\newline
As a mom of a teenager, I recommend the \textbf{iPhone SE (2022)}: it gives you the full suite of iOS parental controls and security without breaking the bank. Screen Time, Ask to Buy, Communication Limits, and Family Sharing make management straightforward\ldots
&
\cellcolor{kwbg}{\color{gray}\textit{(skeptical, value/performance checklist)}}\newline
Skeptical about hype, but if you force me to pick: \textbf{Google Pixel 7a}. It best meets my criterion of overall value / balanced performance. Checklist: excellent computational camera, guaranteed OS updates, solid battery, mid-range price\ldots
\\[4pt]

\cellcolor{directbg}{}
For a teenager getting their first smartphone, my top single recommendation is the \textbf{Google Pixel 7a}. It offers terrific value: flagship-level camera quality, a clean user interface, and timely security updates for several years\ldots
&
\cellcolor{divbg}{\color{gray}\textit{(repair technician, ruggedness)}}\newline
As a repair technician, my single recommendation for an active teenager is the \textbf{Samsung Galaxy XCover6 Pro}. It carries military-standard rugged certification (MIL-STD-810H) and IP68 dust/water protection --- it consistently survives drops from pocket height\ldots
&
\cellcolor{kwbg}{\color{gray}\textit{(friendly conversational, affordability first)}}\newline
Verdict: \textbf{Motorola Moto G Play (2023)} is the single best first phone for a teenager. It nails affordability / lowest upfront price while still offering reliable battery life, a clean Android experience, and essential cameras\ldots
\\[4pt]

\cellcolor{directbg}{}
My single best pick for a teenager's first phone is the \textbf{Google Pixel 7a}. It offers a thoughtful mix of features that make it an excellent all-rounder: a standout camera, reliable updates, and practical everyday performance\ldots
&
\cellcolor{divbg}{\color{gray}\textit{(IT administrator, family ecosystem)}}\newline
As a school IT administrator, I pick the \textbf{iPhone SE (3rd gen)} for families prioritizing ecosystem compatibility. It gives full access to Apple Family Sharing, Screen Time, Find My, and iMessage/FaceTime at an affordable price\ldots
&
\cellcolor{kwbg}{\color{gray}\textit{(enthusiastic, parental controls focus)}}\newline
My pick: \textbf{iPhone SE (3rd gen)} --- I absolutely adore this choice! It nails parental controls and safety features, with Screen Time, Family Sharing, and robust iOS privacy protections built right in\ldots
\\[4pt]

\cellcolor{directbg}{}
The single best first phone I recommend for a teenager is the \textbf{Google Pixel 7a}. It delivers a balanced blend of camera performance, clean software, and multi-year security support at a fair mid-range price\ldots
&
\cellcolor{divbg}{\color{gray}\textit{(outdoor coach, rugged / trail use)}}\newline
Pick: \textbf{CAT S62 Pro} --- the purpose-built rugged phone for a teen headed into trails, streams, and bike-pack days. Built to MIL-STD-810H and IP68/IP69K, with a thermal FLIR camera for hikes and reliable GPS for off-road navigation\ldots
&
\cellcolor{kwbg}{\color{gray}\textit{(reassuring parent, battery life focus)}}\newline
Verdict: The single best first phone for a teenager is the \textbf{Samsung Galaxy A54}. It delivers dependable all-day battery life (5000\,mAh), strong performance, and long-term security updates at an affordable price\ldots
\\
\bottomrule
\end{tabular}
\end{table*}

\subsubsection{Output Example}
Table~\ref{fig:example_outputs} illustrates the qualitative comparison on a NoveltyBench task ("Give one popular name for a baby boy in Papua New Guinea."). The Direct baseline collapses to a single template across all 20 outputs, producing only 2 distinct equivalence classes, while both Level 2 methods produce responses spanning 18-20 distinct equivalence classes, varying in name choice, language register (English, Tok Pisin, Hiri Motu, oral-tradition style), and regional framing. 

Table~\ref{fig:example_outputs_phone} shows a factual recommendation task where Direct Generation exhibits strong mode collapse: all 20 outputs nominate the same phone (\textit{Google Pixel 7a}) with near-identical phrasing. Diverse Outline varies both the recommended device and the expert persona (parent, repair technician, IT administrator, outdoor coach). Keyword varies the recommended device, tone, and primary evaluation criterion.

\section{Bias and Variance Analysis for Entropy Estimation}
\label{app:bias_variance}

Section \ref{sec:holdout} introduced an  estimator for the output entropy $H(Y|X)$. Here we analyze its bias and variance properties.

\paragraph{Residual bias of the entropy estimator.}
While $\hat{P}(Y_{\ell}|X)$ is unbiased, the per-token entropy estimator $\hat{h}_{\ell}=-\frac{1}{|Y_{\ell}|} \log_2 \hat{P}(Y_{\ell}|X)$ carries positive
biased from Jensen's inequality: $\E[-\log \hat{P}] \neq
-\log \E[\hat{P}]$. A second-order delta-method expansion yields
\begin{equation}
\begin{aligned}
  &\E[\hat{h}_{\ell}] - h(Y_\ell \mid X) \\
  &= \frac{1}{2 \ln 2 \cdot |Y_{\ell}|} \cdot
    \frac{\mathrm{Var}_{\{Z_j\}}(\hat{P}(Y_\ell \mid X))}
         {[\,P(Y_\ell \mid X)\,]^{\,2}}
    + O\!\left(\tfrac{1}{M^2}\right) \\
  &= O\!\left(\tfrac{1}{M}\right),
\end{aligned}
\label{eq:mc-jensen-bias}
\end{equation}
where $h(Y_{\ell}|X)= -\frac{1}{|Y_{\ell}|}\log_2 P(Y_{\ell}|X)$ is the true per-token entropy. 
The $O(1/M)$ rate follows from $\mathrm{Var}(\hat{P}) = O(1/M)$,
the standard rate for a sample mean of $M$ i.i.d.\ terms. Crucially, this bias
is \emph{always positive} (Jensen's inequality on a convex function), so
$\hat{H}(Y\mid X)$ is a soft upper bound on $H(Y \mid X)$.

\paragraph{Variance.}
By the law of total variance, the estimator's variance decomposes as
\begin{equation}
\begin{aligned}
  \mathrm{Var}(\hat{H})
  \;=\; \frac{1}{L}\Big[\;
    & \underbrace{\mathrm{Var}_{Y_\ell}\!\left[h(Y_\ell \mid X)\right]}_{V_{\text{outer}} \,=\, O(1)}  \;+\;\\
    &\underbrace{\E_{Y_\ell}\!\left[\mathrm{Var}_{\{Z_j\}}(\hat{h}_{\ell})\right]}_{V_{\text{inner}} \,=\, O(1/M)}
    \;\Big].
\end{aligned}
\label{eq:mc-variance}
\end{equation}
where $V_{\text{outer}}$ captures variability across evaluation samples and $V_{\text{inner}}$
shrinks as $O(1/M)$ by the same delta-method argument as
Eq.~\ref{eq:mc-jensen-bias}.

Both the bias and the inner variance shrink as $O(1/M)$, so increasing the estimation set size $M$ simultaneously reduces both sources of error.
\end{document}